# New Optimization Approach Using Clustering-Based Parallel Genetic Algorithm

**Masoumeh Vali**

Department of Mathematics, Dolatabad Branch, Islamic Azad University, Isfahan, Iran
E-mail: vali.masoumeh@gmail.com

**Abstract**

In many global Optimization Problems, it is required to evaluate a global point (min or max) in large space that calculation effort is very high. In this paper is presented new approach for optimization problem with subdivision labeling method (SLM) but in this method for higher dimensional has high calculation effort. Clustering-Based Parallel Genetic Algorithm (CBPGA) in optimization problems is one of the solutions of this problem. That the initial population is crossing points and subdividing in each step is according to mutation. After labeling all of crossing points, selecting is according to polytope that has complete label. In this method we propose an algorithm, based on parallelization scheme using master-slave. SLM algorithm is implemented by CBPGA and compared the experimental results. The numerical examples and numerical results show that SLMCBPGA is improved speed up and efficiency.

## 1 Introduction

Optimization problems abound in most fields of science, engineering, and technology. In many of these problems it is necessary to compute the global optimum (or a good approximation) of a multivariable function. The variables that define the function to be optimized can be continuous and/or discrete and, in addition, they often have to satisfy certain constraints. Global optimization problems belong to the complexity class of NP-hard problems. Such problems are very difficult to solve, traditional descent optimization algorithms based on local information are inadequate for solving them.

In this paper, we are primarily concerned with the following optimization problems Such as f ($x_1$,$x_2$, ...,$x_m$)Where each $x_i$ is a real parameter object to $a_i \leq x_i \leq b_i$ For some constants $a_i$and $b_i$.we present new method according to subdividing labeling method(SLM) .in this approach , we can earn global points (max/min) with at least step and the simplest algebra operations that it does not require deviation (The following paper will be discussed more) [1].

In many Optimization Problems it is necessary to evaluate a large space for finding global point then the calculation process is very time-consuming. Some of the great disadvantages methods concerning Optimization Problems are dealing with large search spaces and with an extremely high calculation cost.SLM in Optimization Problems that has down complexity time rather than other Optimization Problems, but SLM has high computational in up dimensional and space.

The use of parallelism Genetic Algorithms in Optimization problem are one of the solutions for SLM. There are many studies that are related to parallel genetic algorithms [2, 3, 4, 5, 6, and 7] for searching space method for finding global point. These algorithms allow the use of larger space is searched synchronal; they can use more memory to cope with larger space and dimensions.

Parallel genetic algorithm is one of the developments of genetic algorithms, it is very important to the new generation of intelligent computer. Genetic algorithm has a high degree of parallelism in operation. Many researchers are searching the strategy to implement the genetic algorithms on the parallel machine [8].With the rapid development of the massively parallel computers; Parallel genetic algorithm has become a hot research Spot. The combination of the genetic algorithms and parallel computer, which combines the high-speed of parallel machine and the parallelism of the genetic algorithm, facilitates the research and development of the parallel genetic algorithm [9].
Some more reasons to justify their parallelization are that they reduce the probability of finding suboptimal solutions and they can cooperate in parallel with another search technique. In this work we propose alternate strategies with modified replacement scheme used in parallel SLM Genetic Algorithm with master-slave.
This paper starts with the description of related work in section 2. Section 3 gives the outline of Model and problem definition SLM. In section 4, we have text problems of SLM method. In section 5, we have a discussion about how genetic algorithm can be used to SLM method. In section 6, we present schemata for SLM method with GA. De Jong function definition and test problems of De Jong by SLM GA method are in sections 7 and 8. In section 9 is shown Experimental Results and Summaries of Experimental Results is in section 10**.** The discussion ends with a conclusion and future trend.

## 2 Serial Schema SLM

In this section, we define model of SLM and present serial algorithm for this then Several test problem implement.

### 2.1 Problem Definition for SLM

Supposing F( $x_1, x_2, x_3 \ldots\ldots, x_n$) with constraint$a_i \leq x_i \leq b_i$ , we want to earn global point to this in order to following serial algorithm SLM:

**Step1:** Draw the diagrams for $x_i = b_i \; and \; x_i = a_i$ for i=1, 2,…, n ; then, we find crossing points which equal $n^2$and h= min$\frac{|x_i| + |x_j|}{2}$ for $1 \leq$ i, j $\leq$ n.

**Step2:** Suppose that the point$(a_1, b_1, b_2 \ldots\ldots b_{n-1})$ is one of the crossing points. Consider the values of $\pm h$ gained by step1 and then do the algebra operations on this crossing point as shown in Table 1. Their maximum number in dimensional space is $2 * (\binom{n}{1} + \binom{n}{2} + \binom{n}{3} + \cdots + \binom{n}{n})$ .

Table 1: The number of point is produced by algebra operation on primary point.

| |
|---|
| $(\boldsymbol{a_1 \pm h, b_1, b_2 \ldots\ldots b_{n-1}})$ |
| $(\boldsymbol{a_1, b_1 \pm h, b_2 \ldots\ldots b_{n-1}})$ |
| ⋮ |
| $(\boldsymbol{a_1, b_1, b_2 \ldots\ldots b_{n-1} \pm h})$ |
| ⋮ |
| $(\boldsymbol{a_1 \pm h, b_1 \pm h, b_2 \ldots\ldots b_{n-1}})$ |
| ⋮ |
| $(\boldsymbol{a_1 \pm h, b_1 \pm h, b_2 \pm h, \ldots, b_{n-1} \pm h})$ |

**Step 3:** The function value is calculated for all points of step 2 and the value of these functions is compared with f$(a_1, b_1, b_2 \ldots\ldots b_{n-1})$ . At the end, we will select the point which has the minimum value and we will call it$(c_1, c_2, c_3 \ldots\ldots c_n)$.

**Note:** If we want to find the optimal global max, we should select the maximum value.

**Step 4:** In this step, the equation1is calculated:

$$(c_1, c_2, c_3 \ldots\ldots c_n) - (a_1, b_1, b_2 \ldots\ldots b_{n-1}) = (d_1, d_2, d_3 \ldots\ldots d_n) \qquad (1)$$

**Step 5:** According to the result of step 4, the point $(a_1, b_1, b_2 \ldots\ldots b_{n-1})$ is labeled according to equation (2).

$$l(a_1, b_1, b_2 \ldots\ldots b_{n-1}) = \begin{cases} 0 & d_1 \geq 0, \ldots, d_n \geq 0 \\ 1 & d_1 < 0, d_2 \geq 0, \ldots, d_n \geq 0 \\ 2 & d_2 < 0, d_3 \geq 0, \ldots, d_n \geq 0 \\ \vdots & \\ N & d_n < 0 \end{cases} \qquad (2)$$

**Step 6:** Go to step 7 if all the crossing points were labeled, otherwise repeat the steps 2 to 5.

**Step 7:** In this step, the complete labeling polytope is focused. In fact, a polytope will be chosen that has complete labeling in different dimensions as shown in Table 2.

Table 2: complete labeling with different dimensional

| Dimension | Complete Label |
|---|---|
| 2 | 0,1,2 |
| 3 | 0,1,2,3, |
| | |
| n | 0,1,2,3,…,n |

**Step 8:** In this step, all sides of the selected polytope (from step 7) are divided into 2 according to equation 3 and we repeat steps 3 to 7for new crossing points.

$$h = \min\left\{\frac{|x_i| + |x_j|}{2}\right\} \quad 1 \leq i, j \leq n \qquad (3)$$

**Step 9:** Steps 2 to 8 are repeated to the extent that h → 0and the result is global min or max point.

### 2.2 Test Problems of SLM:

In this part, we have three examples of SLM.

#### 2.2.1 Test Problem 1

This function is a continuous function .The optimization problem is equation 4.

$$min f(x_1, x_2) = {x_1}^2 + (x_2 - 0.4)^2 \qquad -2 < x_i < 2, i = 1,2 \qquad (4)$$

The function achieves the minimum when $x_1 = 0$, and $x_2 = 0.4$.in this example,$h_i \in \{4, 2, 1, 0.5, 0.25\}$ mutation probability$p_m = 1$. The completely label square obtains through the iteration, the search scope for both $x_1$ , $x_2$ are (-2,2) , (0,2) , (0,1) , and finally (0, 0.5) respectively (as show from Figure 1 to Figure 3 and from table 3 to 5 ). During iterations, squares are contracting to (0, 0.5) gradually, if we started from$h_1 = 1$, we got closer answer i.e. (0, 0.4).

Figure 1: Initial Population of $f_1$

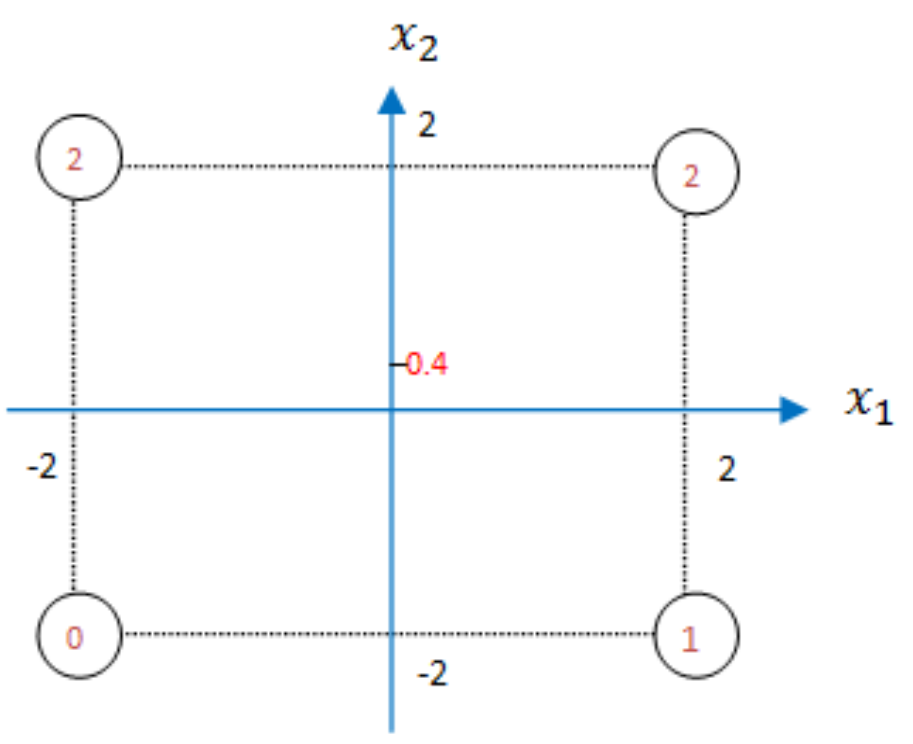

| $h_1$=4, P(0): | $h_2$=2 | P(1): | $l(x)$ | Solution |
|---|---|---|---|---|
| (-2,2) | | (0,0) | 2 | (0,0) |
| (2,2) | | (0,0) | 2 | |
| (-2,-2) | | (0,0) | 0 | |
| (2,-2) | | (0,0) | 1 | |

Table 3: Initial Population of $f_1$

Figure 2: First generation of $f_1$

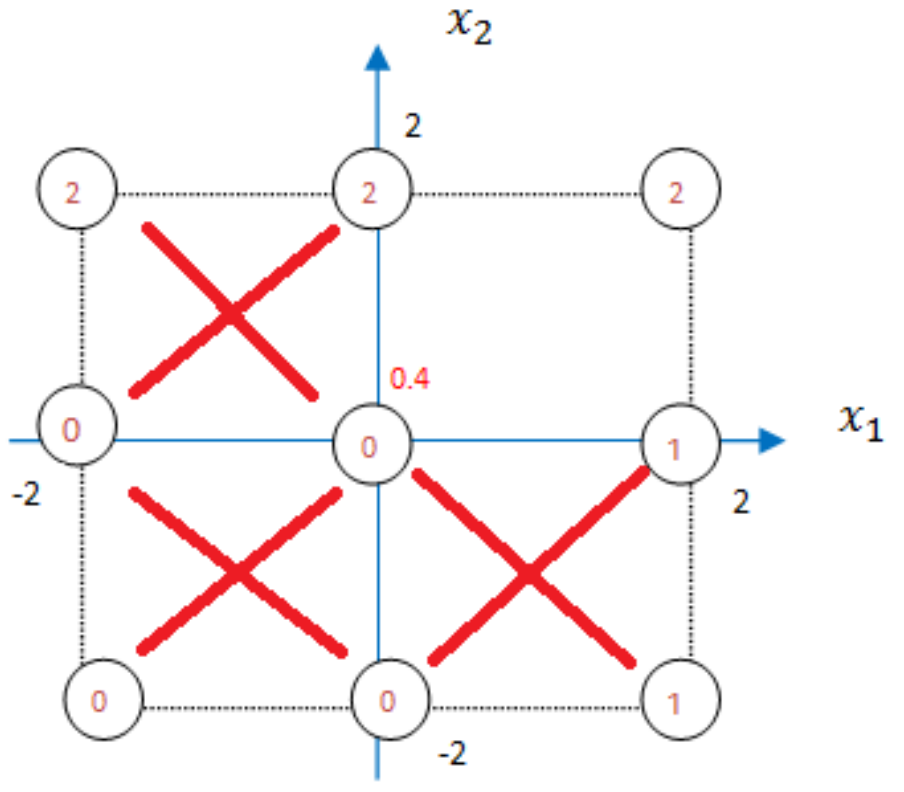

| $h_2$=2 P(1): | $h_3$=1 | P(2): | $l(x)$ | Solution |
|---|---|---|---|---|
| (-2,2) | | (-1,1) | 2 | (0,0) |
| (2,2) | | (1,1) | 2 | |
| (-2,-2) | | (-1,- | 0 | |
| (2,-2) | | (1,-1) | 1 | |
| (2,0) | | (1,0) | 1 | |
| (0,2) | | (0,1) | 2 | |
| (-2,0) | | (-1,0) | 0 | |
| (0,-2) | | (0,-1) | 0 | |

Table 4: First generation of $f_1$

Figure 3: Second generation of $f_1$

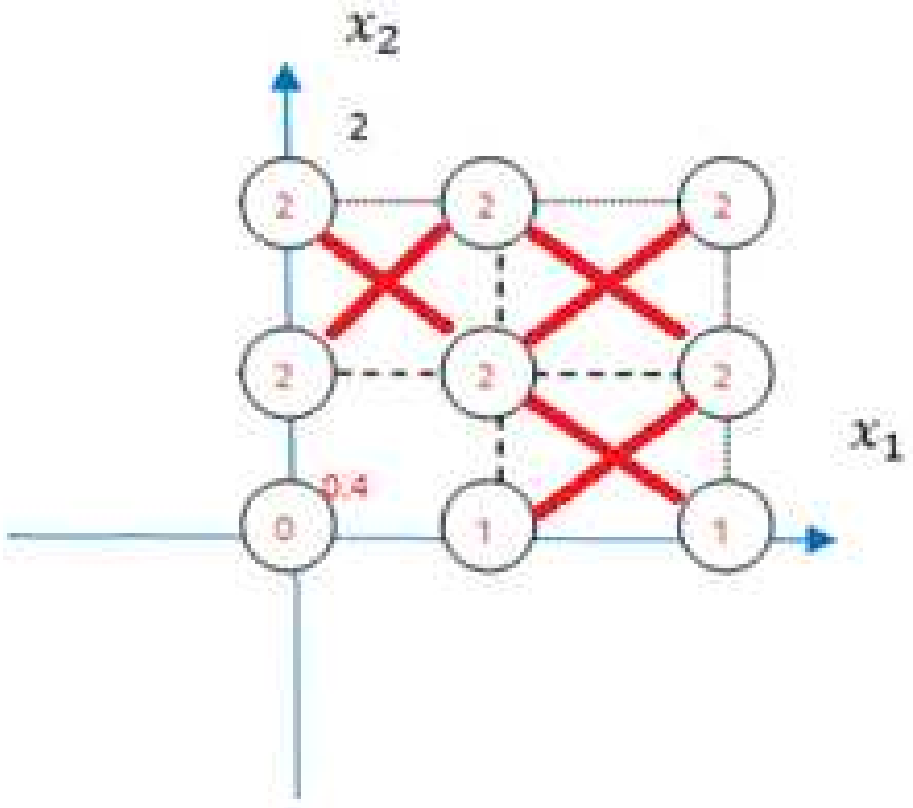

| $h_3$=1 P(2): | $h_4$=.5 | P(3): | $l(x)$ | Fixed point |
|---|---|---|---|---|
| (-1,1) | | (-0.5,0.5) | 2 | |
| (1,1) | | (0.5,0.5) | 2 | |
| (-1,-1) | | (-0.5,-0.5) | 0 | |
| (1,-1) | | (0.5,-0.5) | 1 | |
| (1,0) | | (0.5,0.5) | 1 | |
| (0,1) | | (0,0.5) | 2 | |
| (-1,0) | | (-0.5,0.5) | 0 | **(0,0.5)** |
| (0,-1) | | (0,-0.5) | 0 | |
| (0,0) | | (0,0.5) | 0 | |
| (-1,2) | | (-0.5,1.5) | 2 | |
| (2,2) | | (1.5,1.5) | 2 | |
| (-2,1) | | (-1.5,0.5) | 2 | |
| | | | | |

Table 5: Second generation of $f_1$

We compare SLM of test problem 1 with two differential methods and random search that results are shown in table 7. Results show SLM earn global optimal same as other methods.

Table 6: comparison between test problem #1 and other three methods

| **Algorithms** | Iteration | Optimal point | Best Point | Standard deviation |
|---|---|---|---|---|
| **SLM** | 6 | (0,0.4375) | (0,0.4) | (0, 0.0375) |
| **RS** | 1000 | (0,0) | | (0,0.4) |
| **RSW**($\boldsymbol{x}^{\textbf{initial}} = (\mathbf{14.0356}, \mathbf{14.0356})$) | 500 | ( 0, 0. 49999996 ) | | (0,0.09999996) |
| **SA** | 150 | (0,0.42) | | (0,0.02) |

### 2.2.2Test problem2

In this problem, we choose a nonlinear optimization problem with two continuous variables.

(5)

This multimodal function has many local optimal in its domain. The GA keeps each local and global optimal one found in squares labeled completely. In this example, for $h_i \in \{6, 3, 1.5\}$ while mutation probability $p_m = 1$, as shown in figure 6, these points can be gotten. Three following generation have been shown in the first quarter of the coordinates system (see Figure 4 to Figure 6).

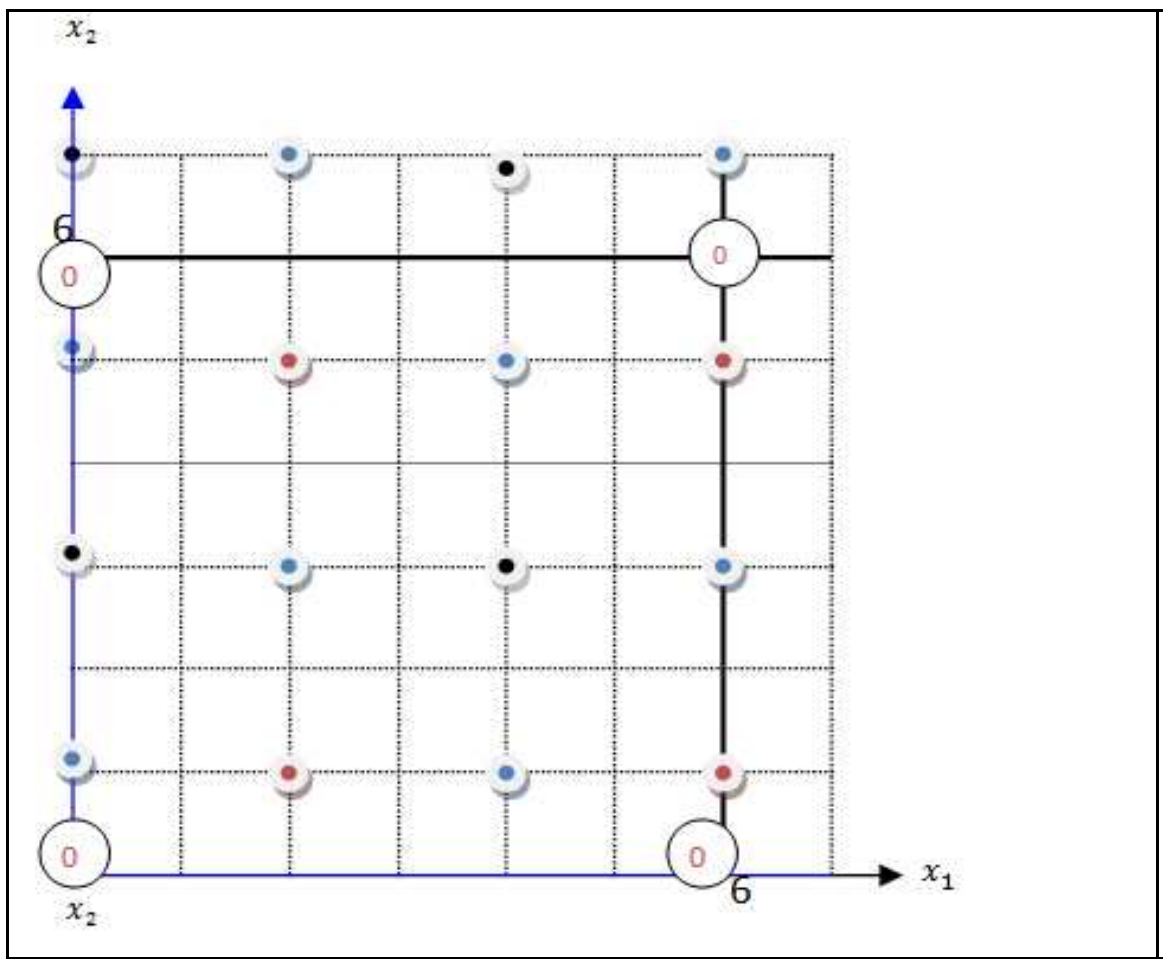

Figure 4: First generation of $f_2$

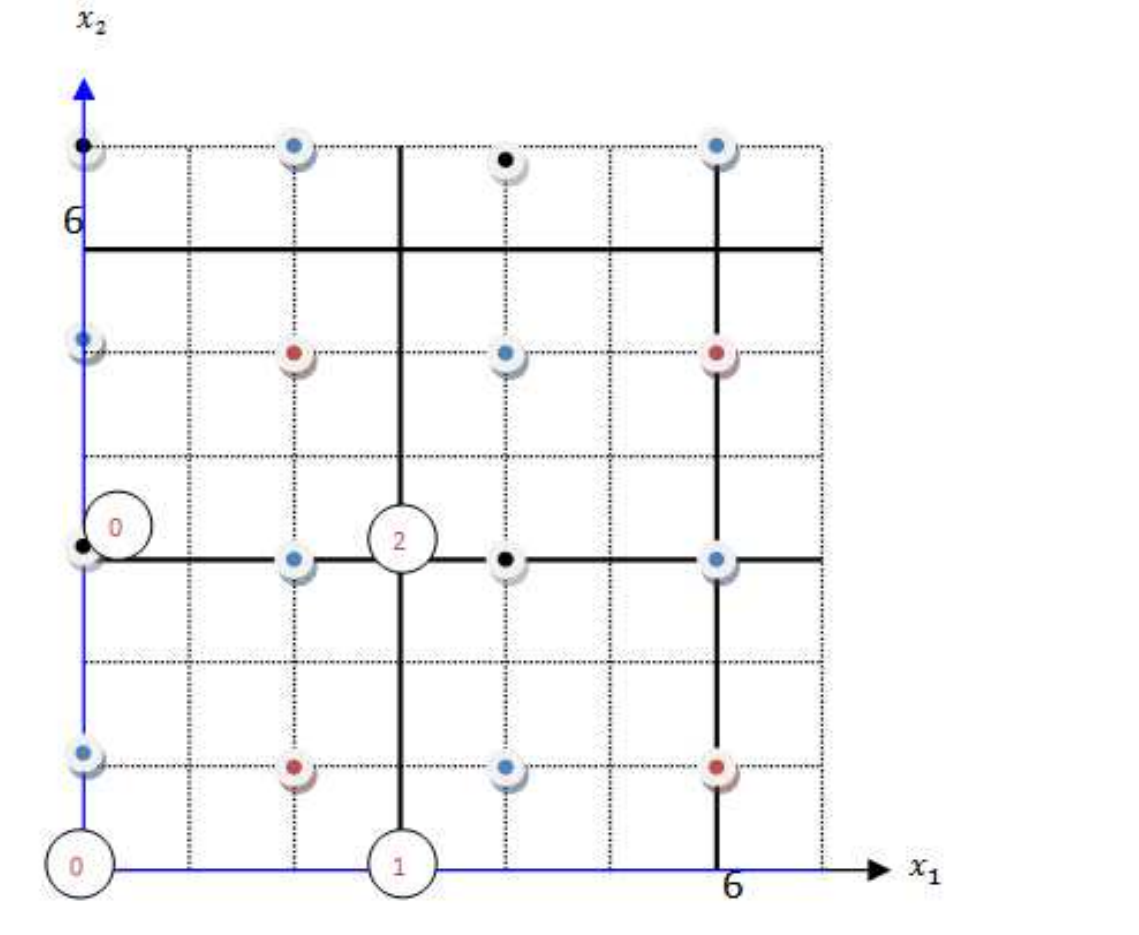

Figure 5: Second generation of $f_3$

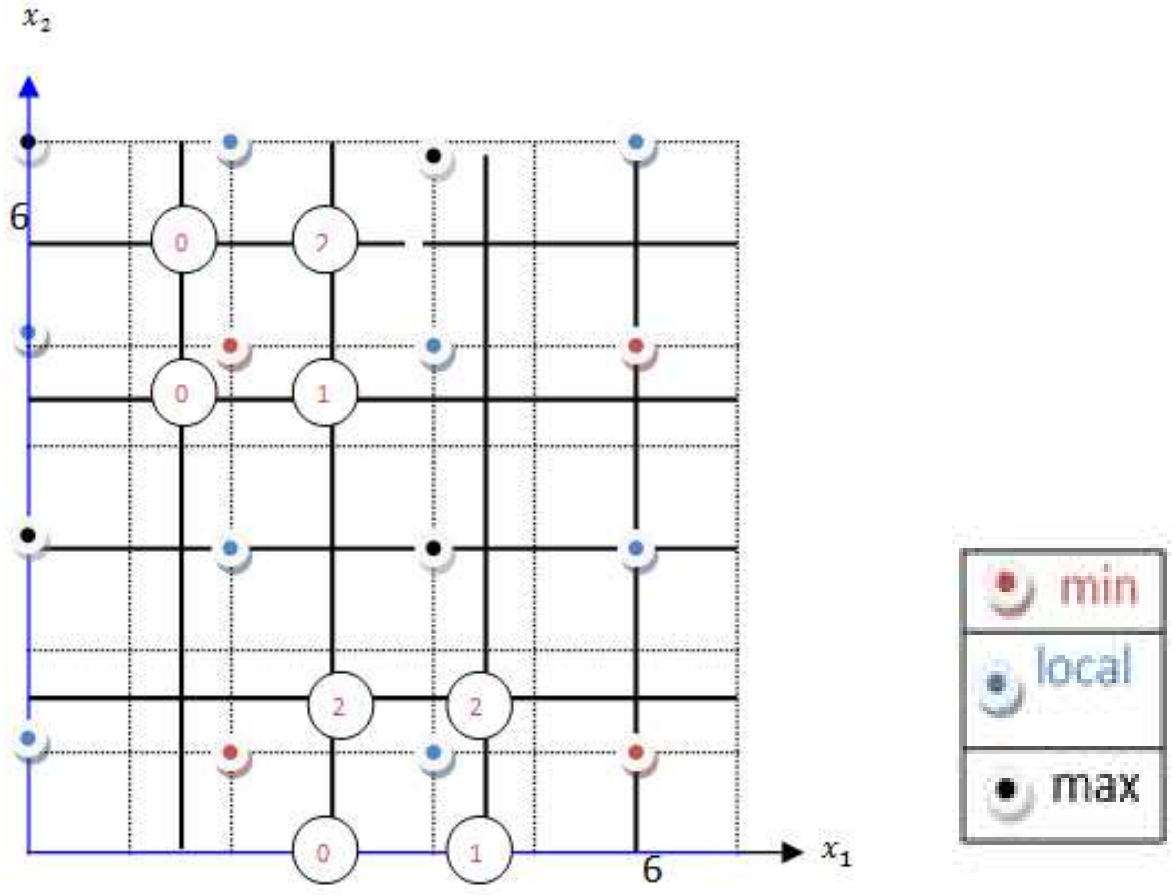

Figure 6: Third generation of $f_2$

Table 8: comparison between test problem #2 and other three methods

| Algorithms | Iteration | Optimal point | Best Point | Standard deviation |
|---|---|---|---|---|
| **SLM** | 7 | $(\pm 4k_1 \pm 1.78125$, $2k_2$+0.78125) ,$k_1$=0,1, $k_2$=0,2 | ($\pm 4k_1 \pm 2$, $2k_2$+1), $k_1$=0,1, $k_2$=0,2 | (0.21875, 0.21875) |
| **RS** | 1000 | ($\pm 4k_1 \pm 2.48$, $2k_2$+1.25), $k_1$=0,1, $k_2$=0,2 | | (0.48, 0.25) |
| **RSW**$(\boldsymbol{x}^{\text{initial}} = (\mathbf{18.048}, \mathbf{14.899}))$ | 500 | ($\pm 4k_1 \pm 1.999$,$2k_2$+.0999), $k_1$=0,1, $k_2$=0,2 | | (0.001,0.001) |
| **SA** | 150 | ($\pm 4k_1 \pm 1.9499$,$2k_2$+.0999) $k_1$=0,1, $k_2$=0,2 | | (0.0501,0.001) |

We compare SLM of test problem 2 with two differential methods and random search that results are shown in table 8. Results show SLM earn global optimal same as other methods.

### 2.2.3 Test problem3

The Easom function [Eas90] is a unimodal test function, where the global minimum has a small area relative to the search space. The function was inverted for minimization.

Function definition:

$$f_{\text{Easo}}(x_1, x_2) = -\cos(x_1).\cos(x_2).e^{-\left((x_1-\pi)^2+(x_2-\pi)^2\right)}; \qquad -100 \le x_i \le 100, i = 1{,}2$$

global minimum:

$$f(x_1, x_2) = -1; \quad (x_1, x_2) = (\pi_i, \pi_i)$$

Figure 7: Initial Population of $f_{\text{Easo}}$

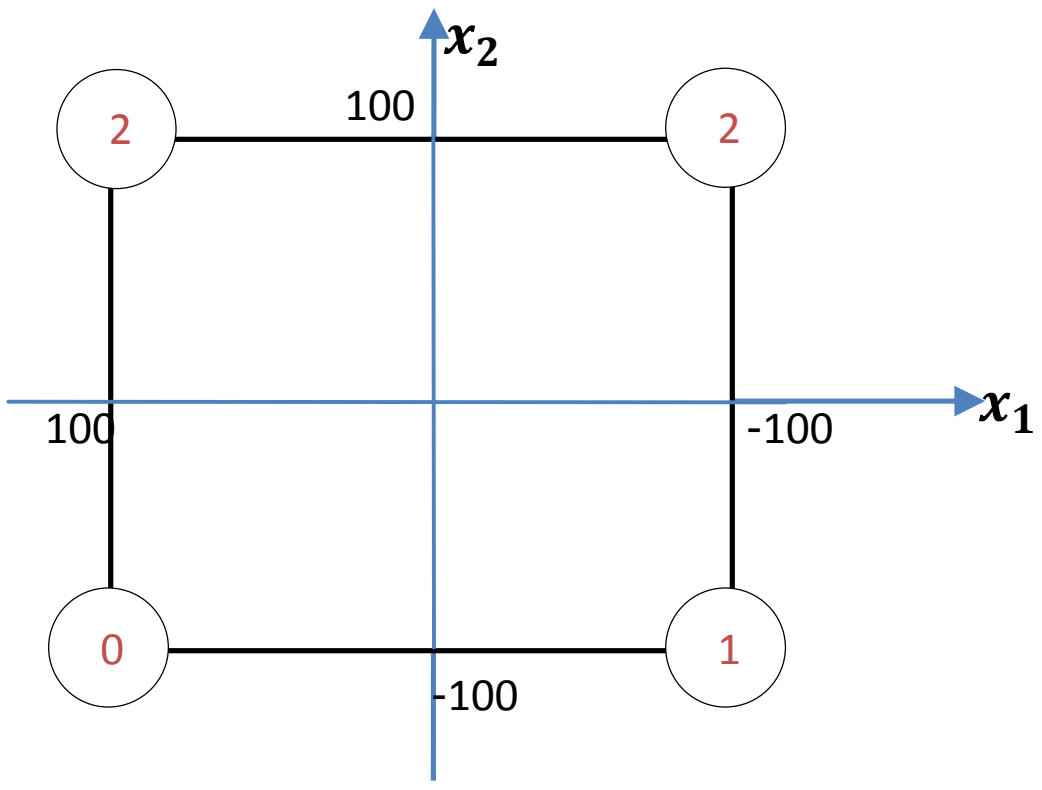

| $h_1$=200, P(0): | $h_2$=100 | P(1): | $l(x)$ | Solution |
|---|---|---|---|---|
| (-100,-100) | $\overset{M}{\rightarrow}$ | (0,0) | 0 | (0,0) |
| (100,100) | | (0,0) | 2 | |
| (-100,100) | | (0,0) | 2 | |
| (100,-100) | | (0,0) | 1 | |

Table 9: Initial Population of $f_{\text{Easo}}$

Figure 8: First generation of $f_{\text{Easo}}$

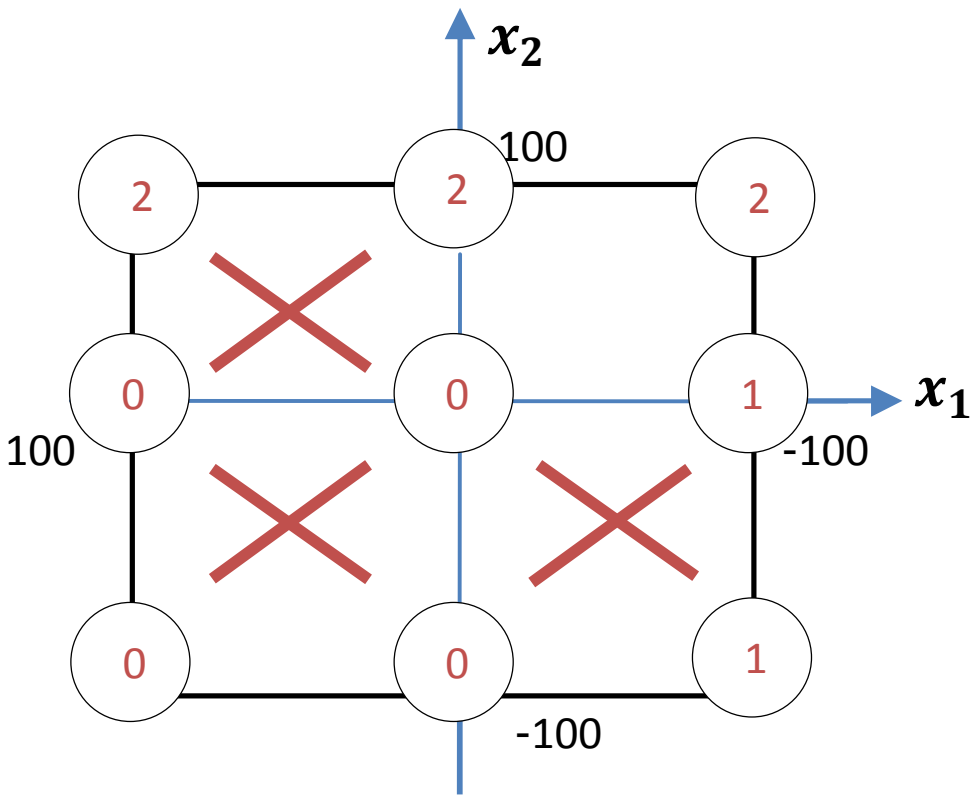

| $h_2$=100 P(1): | $h_3$ =50 | P(2): | $l(x)$ | Solution |
|---|---|---|---|---|
| (-100,0) | $\overset{M}{\rightarrow}$ | (-50,0) | 0 | (0,0) |
| (0,-100) | | (0,-50) | 0 | |
| (100,0) | | (50,0) | 1 | |
| (0,100) | | (0,50) | 2 | |
| (0,0) | | (0,0) | 0 | |

Table 10: First generation of $f_{\text{Easo}}$

Figure 9: Second generation of $f_{\text{Easo}}$

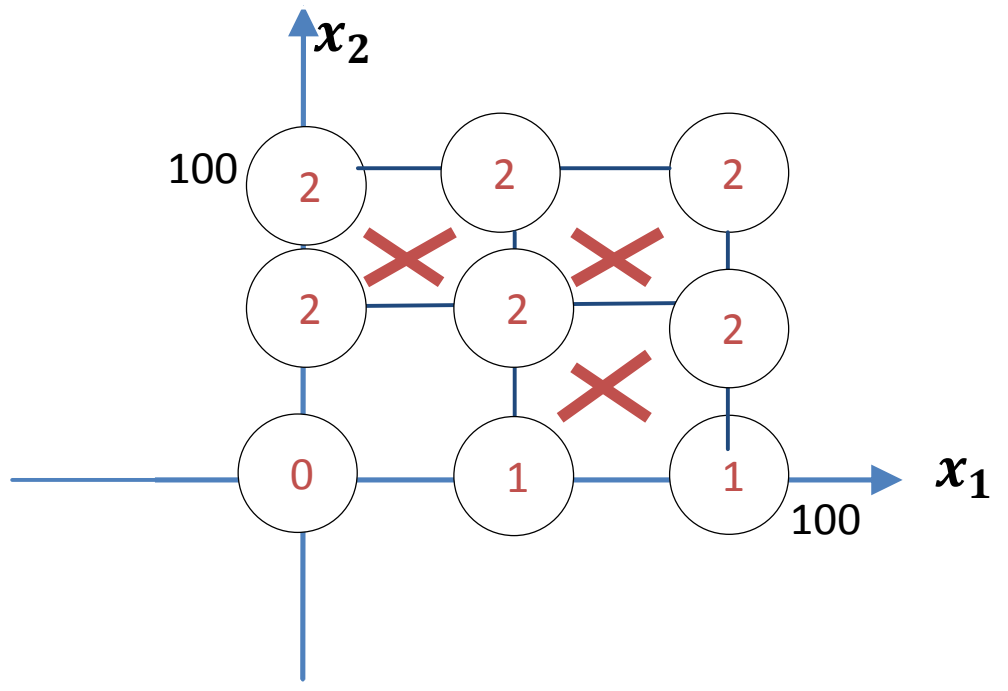

| $h_3$=50, P(2): | $h_4$ =25 | P(3): | $l(x)$ | Solution |
|---|---|---|---|---|
| (0,50) | $\xrightarrow{M}$ | (0,25) | 2 | (0,0) |
| (50,0) | | (25,0) | 1 | |
| (50,50) | | (25,25) | 2 | |
| (50,100) | | (25,75) | 2 | |
| (100,50) | | (75,25) | 2 | |

Table 10: Second generation of $f_{\text{Easo}}$

Table11: comparison between test problem #3 and other three methods

| Algorithms | **Iteration** | **Optimal point** | **Best point** | **Standard deviation** |
|---|---|---|---|---|
| SLM | 11 | (3.3203125,3.3203125) | $(\pi_i, \pi_i)$ | (0.1803125,0.1803125) |
| RS | 1500 | (3.99987, 3.9876988) | | (0.85987, 0.8476988) |
| RSW$(x^{\text{initial}} = (\mathbf{1.048, 0.89}))$ | 700 | (3,3) | | (.014,0.14) |
| SA | 1200 | (3,3) | | (.014,0.14) |

We compare SLM of test problem 3 with two differential methods and random search that results are shown in table 8. Results show SLM earn global optimal same as other methods.

# 3 SLM CLUSTERING-BASED PARALLEL GENETIC ALGORITHM (SCBPGA)

## 3.1 The SLM Parallel Algorithm (SLMPA)

The parallel computing [10] is following. In the parallel computer, an application is divided into multiple sub-tasks, these tasks are assigned to different processors, and then each processor cooperates with each other and parallel completes the tasks. So as to speed up solving the task or increase the size of the application of the purpose of solving. Simplify speaking, it can be seen as supercomputing in the parallel computer or distributed computer and other high performance computer system. The parallel computer system at least includes two or more the processors .They is connected for data communication to achieve parallel processing of the processors. [11]
The SLM by dividing search space in each step, our search space became more and in each search space are lots of points so each search space is allocated to one processor. If the number of processors is low then several of search space is allocated to one processor.

In this article is combined Parallel Computing with GA that will be discussed about GA and PGA in section 3.2 and section 3.3.

## 3.2 Genetic Algorithm SLM

In this section, we discussed about initialization, selection, mutation and schema for SLM GA. At last, we present several test problems that are implemented with SLM GA.

### 3.2.1 Initialization

Initially many individual solutions are generated to form an initial population. The population size depends on the nature of the problem , the initial population in SGA is generated accordingly to dimension for example initial population for 2- dimensional is four, for 3- dimensional is eight and so on for n- dimensional is $2^n$according to table 10 initial population in this method earn through crossing point with equation$2^n$ .In genetic algorithm like other evolutionary algorithm, its optimal solutions are points that the algorithm improves keeps or returns to them after a certain number of iterations because these points meet required criteria of the algorithm.

| Dimensional | Crossing points |
|---|---|
| 2 | $2^2$ |
| 3 | $2^3$ |
| 4 | $2^4$ |
| ⋮ | ⋮ |
| n | $2^n$ |

Table 10: The initial population in SLM GA

### 3.2.2 Selection

During each successive generation, a proportion of the existing population is selected to breed a new generation. Individual solutions are selected through a fitness-based process, where fitter solutions (as measured by a fitness function) are typically more likely to be selected. Certain selection methods rate the fitness of each solution and preferentially select the best solutions.

Supposing that algorithm is searching a point x which can make continuous function of $f(x)$to achieve its optimal point. The necessary and sufficient condition of extreme point is that this point gradient is 0 that is $\nabla f(x) = 0$.
For self-mapping $g:\mathbb{R}^n \to \mathbb{R}^n$, we say $x \in \mathbb{R}^n$is a fixed point of $g$ if $g(x) = x$ , then we can convert the solution of zero point problems to fixed point ones of function $g(x) = x + \nabla f(x)$.

Supposing that definition domain of $f(x_1, x_2, \dots, x_n)$ is that $a_1 \leq x_1 \leq a_2$ $, a_3 \leq x_2 \leq a_4 , \dots, a_r \leq x_n \leq a_s$and dividing the domain into many polytopes with two groups of straight lines of $\{x_1 = mh_i\}, \{x_2 = mh_i\}, \dots, \{x_n = mh_i\}$ in which m is a not negative integer and$h_i$ is a positive quantity relating to precision of the problem. As a result, we can code each point of intersection as $x_1 = a_1 + k_1 h_i$, $x_2 = a_3 + k_2 h_i$,…,$x_n = a_r + k_n h_i$ where $k_1, k_2, \dots, k_n$ are not negative integers, so$(k_1, k_2, \dots, k_n)$ is called the relative coordinates of points. Consequently, by changing $k_1, k_2, \dots, k_n$ relative coordinates of each point in search space is determined.
Attentive points that are pointed above, labeling is done according equation 6:

$$l(x) = \begin{cases} 0, & g_1(x) - x_1 \geq 0, \dots, g_n(x) - x_n \geq 0 \\ 1, & g_1(x) - x_1 < 0, g_2(x) - x_2 \geq 0, \dots, g_n(x) - x_n \geq 0 \\ 2, & g_2(x) - x_2 < 0, g_3(x) - x_3 \geq 0, \dots, g_n(x) - x_n \geq 0 \\ \vdots & \\ n, & g_n(x) - x_n < 0 \end{cases} \qquad (6)$$

The polytope with all different kinds of integer label is called a completely labeled unite, when $h_i \to 0$ within iteration stages, vertices of that polytope approximately converge to one point which is a fixed point.

### 3.2.3 Mutation Operator

For each point coded $(k_1, k_2, \dots, k_n)$, the GA is trying to improve it to reach optimal solution by mutation operator searching all points surrounding it in certain step determined by $h_{i+1}$.

For instance, $(k_1, k_2, \dots, k_n)$ in P (0), initial population, addressing $(x_1 + k_1 h_i, x_2 + k_2 h_i, \dots, x_n + k_n h_i)$will be changed as$(x_1 + \alpha_1, x_2 + \alpha_2, \dots, x_n + \alpha_n)$,$\alpha_1, \alpha_2, \dots, \alpha_n \in \{0, \pm h_{i+1}\}$. Subsequently, the algorithm saves the best-mutated individual among all possible offspring. Therefore, this operator produces new population located on intersection of the next grid. Because of this, coming polytopes are specified to evaluate and label. Furthermore, the next generation is producing from the previous one.

### 3.2.4 Schemata Analysis for SLM Genetic Algorithms

Figure 7: Improvement genetic algorithm flow chart

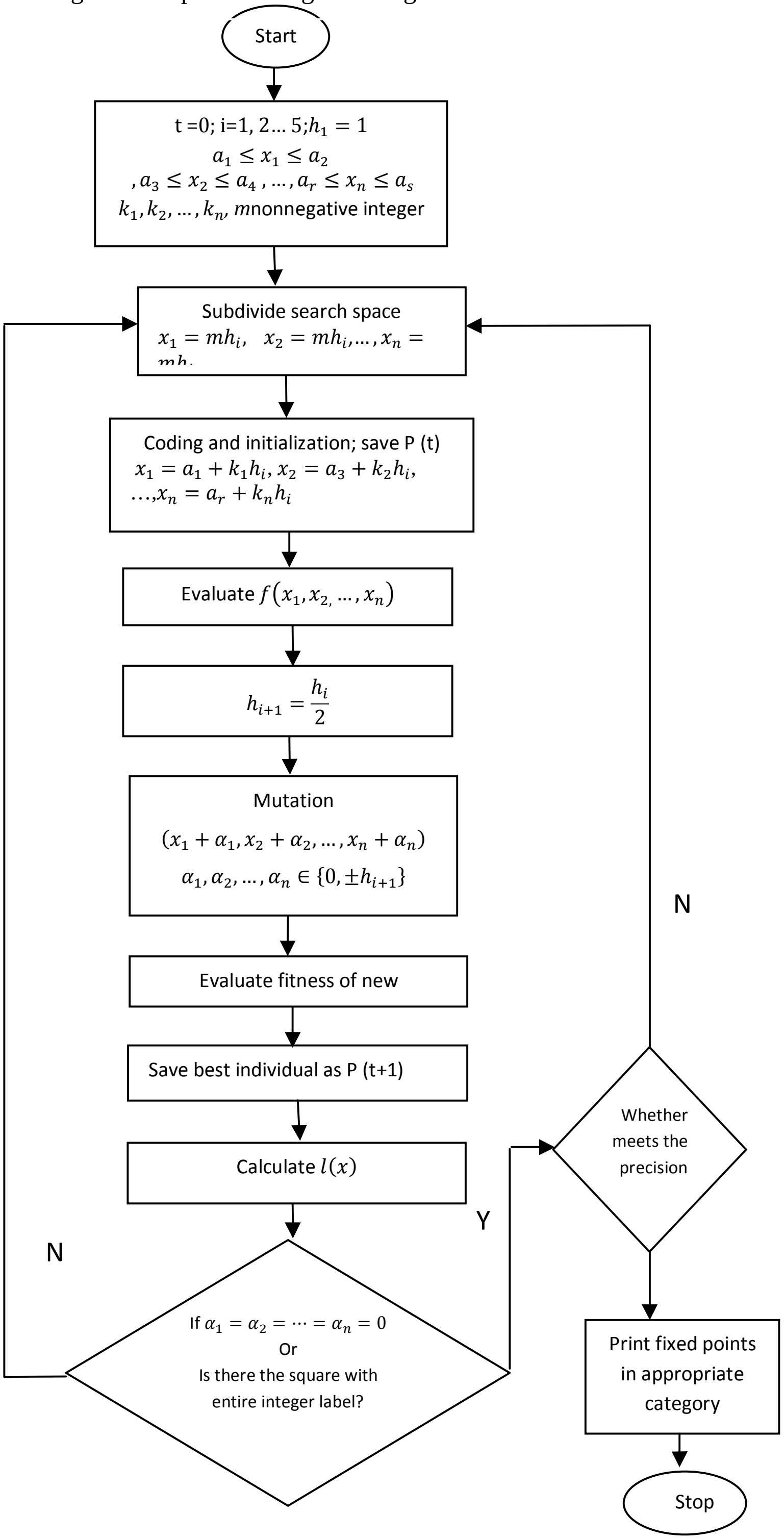

This improved algorithm (Figure 7) makes grid in given scope and encodes each intersection by integer while it starts from the lowest point of the domain. After calculating fitness of each point, it generates the best offspring and computes integer label of the last population for every square. When it found the square labeled completely, subdivides them in order to seek the solution closely. As following, we demonstrate the performance of the improved algorithm by different examples and show how it can categorize fixed points.

### 3.2.5 Test problems SLMGA:

In this section, we present two test problems with SLMGA.

#### 3.2.5.1 Test problem 1

This problem is second De Jong function (Rosen rock's saddle) [19].

$$f_2(x_0, x_1) = 100.(x_0{}^2 - x_1) + (1 - x_0)^2; \qquad x_j \in [-5.12, 5.12] \tag{12}$$

Although $f_2(x_0, x_1)$has just two parameters, it has the reputation of being a difficult minimization problem. The minimum is $f_2(1,1) = 0$. in this function, figures from 8 to 11 and tables from 7 to 10 are shown.

Figure 8: Figure of Rosen rock's saddle

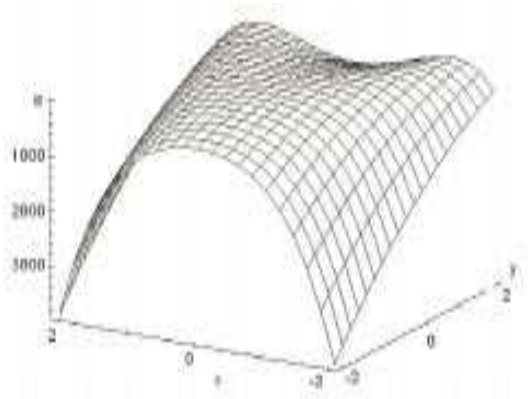

Figure 9: Initial Population of $f_2$

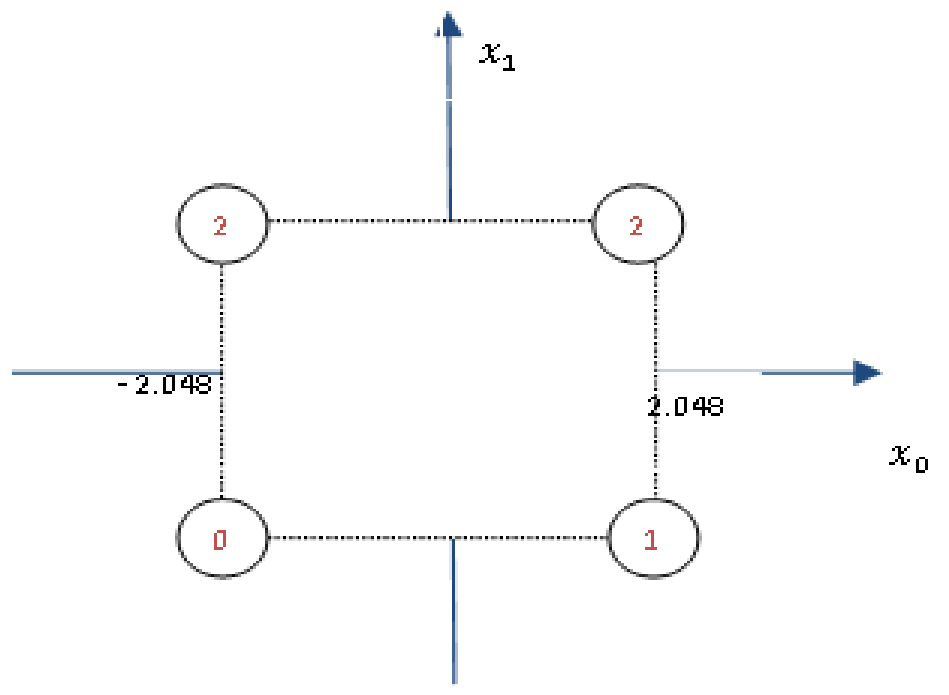

| $h_1 = 4.096$ P(0): | $h_2 = 2.048$ | P(1): | $l(x)$ | Solution |
|---|---|---|---|---|
| (2.048,2.048) | $\overrightarrow{M}$ | (0,0) | 2 | (0,0) |
| (2.048,-2.048) | | (0,0) | 1 | |
| (-2.048,-2.048) | | (0,0) | 0 | |
| (-2.048,2.048) | | (0,0) | 2 | |

Table 7: Initial Population of $f_2$

Figure 10: The First generation of $f_2$

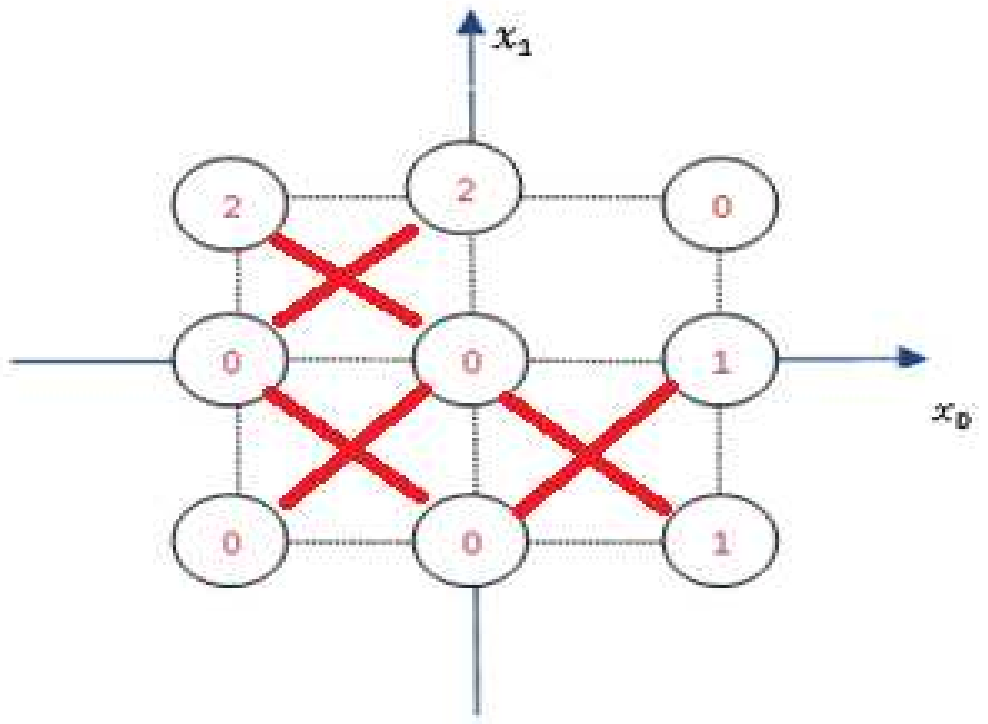

| $h_2 = 2.048$ P(1): | $h_3 = 1.024$ | P(2): | $l(x)$ | Solution |
|---|---|---|---|---|
| (0, 2.048) | $\xrightarrow{M}$ | (-1.024,1.024) | 2 | (1.024,1.024) |
| (0,0) | | (0,0) | 0 | |
| (2.048,0) | | (1.024,1.024) | 1 | |
| (2.048, 2.048) | | (1.024,1.024) | 0 | |

Table 8: The First generation of $f_2$

Figure 11: The second generation of $f_2$

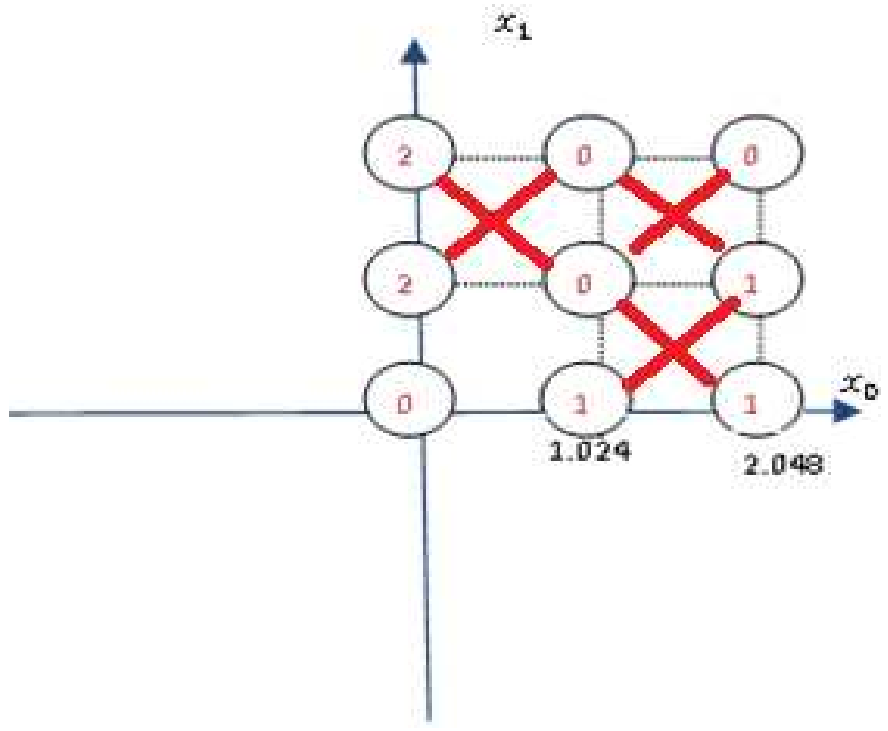

| $h_3 = 1.024$ P(2): | $h_4 = 0.512$ | P(3): | $l(x)$ | Solution |
|---|---|---|---|---|
| (1.024,2.048) | $\xrightarrow{M}$ | (1.536,2.048) | 0 | (1.024,1.024) |
| (2.048,1.024) | | (1.536,1.536) | 1 | |
| (1.024,1.024) | | (1.024,1.024) | 0 | |
| (1.024,0) | | (0.512,0) | 1 | |
| (0,1.024) | | (-0.512,0.512) | 2 | |

Table 9: The Second generation of $f_2$

Figure 12: The Third generation of $f_2$

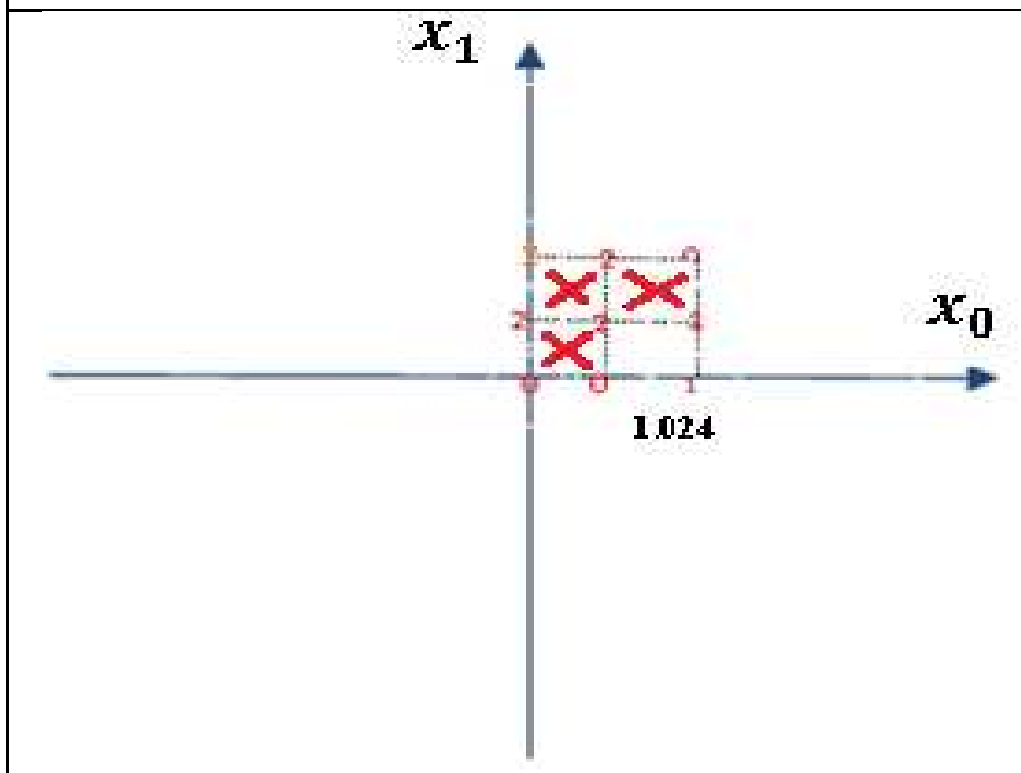

| $h_4 = 0.512$ P(3): | $h_5 = 0.2$ | P(4): | $l(x$ | Fixed point |
|---|---|---|---|---|
| (1.024,0.512) | $\xrightarrow{M}$ | (0.768,0.512) | 1 | (1.024,1.024) |
| (0.512,0) | | (0.768,0.512) | 0 | |
| (0,0.512) | | (-0.256,0.256) | 2 | |
| (0.512,0.512) | | (0.512,0.256) | 2 | |
| (0.512,1.024) | | (0.512,0.768) | 2 | |

Table 10: The Third generation of $f_2$

### 3.2.5.2 Test problem 2

This is test problem of Fifth De Jong function (Shekel's Foxholes).

$$f_5(x_0, x_1) = \frac{1}{0.002 + \sum_{i=0}^{24} \frac{1}{i + \sum_{j=0}^{1} \left(x_j - a_{ij}\right)^6}}; \quad x_j \in [-65.536, 65.536] \qquad (13)$$

With $a_{i0}$=$\{-32, -16, 0, 16, 32\}$ $for\ i = 0,1,2,3,4$ and $a_{i0} = a_{i\ mod\ 5,0}$

As well as $a_{i1}$=$\{-32, -16, 0, 16, 32\}$ $for\ i = 0,5,10,15,20$ and $a_{i1} = a_{i+k,1}$, $k = 1,2,3,4$.

The global minimum for this function is $f_5(-32, -32) \cong 0.998004$.

Figure 13: Figure of Shekel's Foxholes

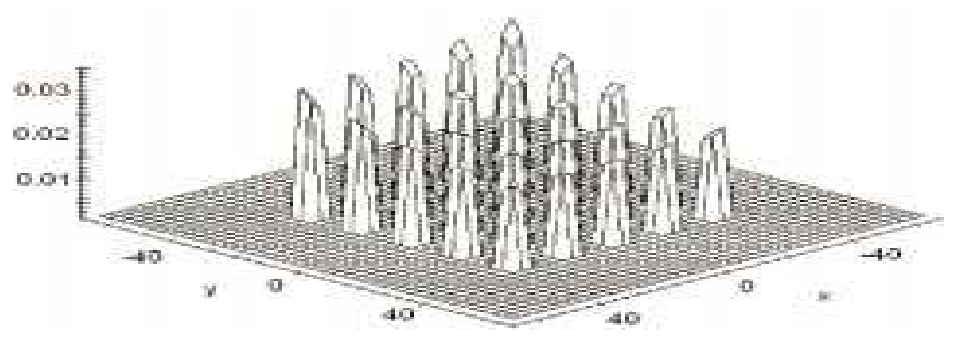

Figure 14: The initial population $f_5$

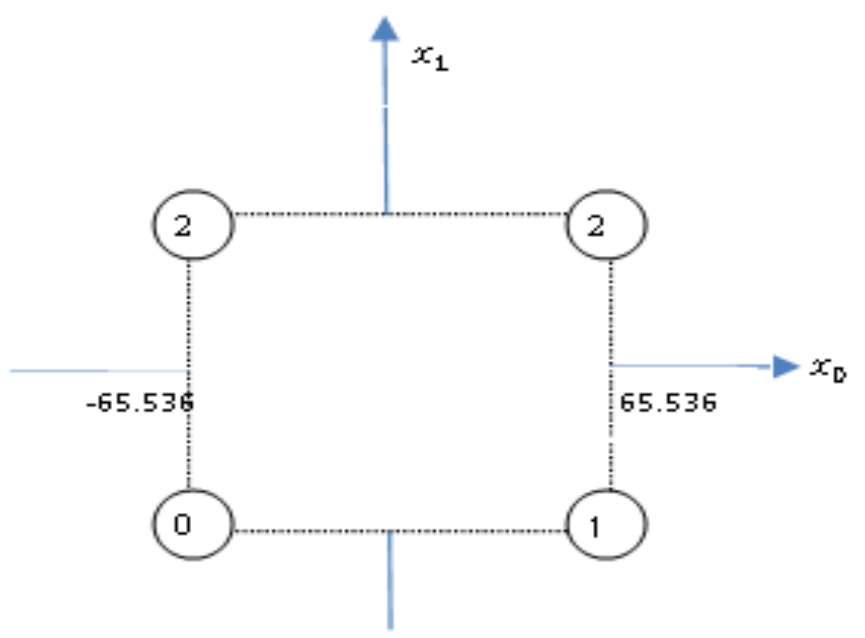

| $h_1 = 131.072$ P(0): | $h_2 = 65.536$ | P(1): | $l(x)$ | Solution |
|---|---|---|---|---|
| (65.536,65.536) | | (0,0) | 2 | |
| (65.536,-65.536) | | (0,0) | 1 | |
| (-65.536, 65.536) | $\xrightarrow{M}$ | (0,0) | 2 | (0,0) |
| (-65.536, -65.536) | | (0,0) | 0 | |

Table 11: The initial population $f_5$

Figure 15: the first generation of $f_5$

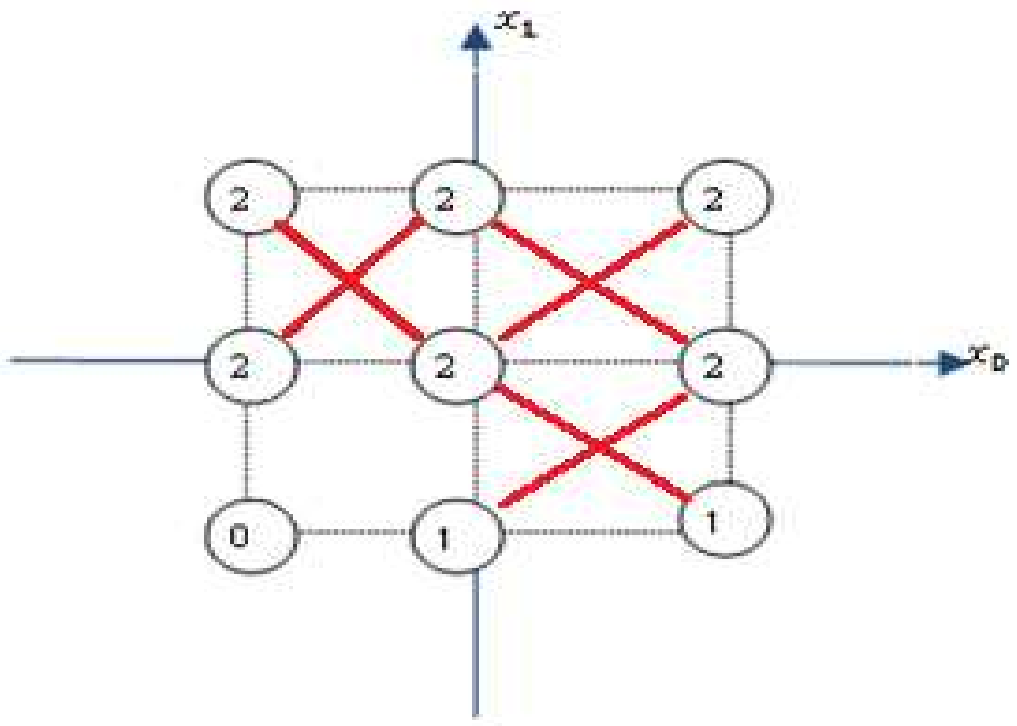

| $h_2 = 65.536$ P(1): | $h_3$ =32.768 | P(2): | $l(x)$ | Solution |
|---|---|---|---|---|
| (65.536,0) | | (32.768,-32.768) | 2 | |
| (0,65.536) | | (-32.768, 32.768) | 2 | |
| (0,0) | $\xrightarrow{M}$ | (-32.768, -32.768) | 2 | (-32.768, -32.768) |
| (-65.536, 0) | | (-32.768, -32.768) | 2 | |
| (0, -65.536) | | (-32.768, -32.768) | 1 | |

Table 12: The first generation of $f_5$

Figure 16: The second generation of $f_5$

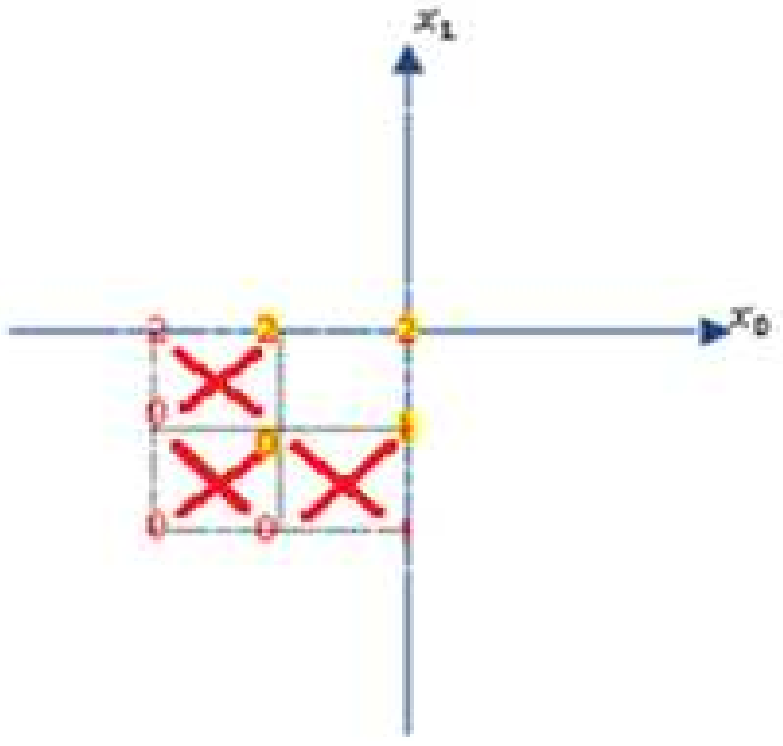

| $h_3$ =32.768 P(2): | $h_4$ =16.384 | P(3): | $l(x)$ | Fixed point |
|---|---|---|---|---|
| (-32.768,0) | | (-32.768,-16.384) | 2 | |
| (-32.768,-32.768) | | (-32.768,-32.768) | 0 | |
| (-32.768,-65.536) | $\xrightarrow{M}$ | (-32.768,-49.152) | 0 | (-32.768,32.768) |
| (0,-32.768) | | (-16.384,-32.768) | 1 | |
| (-65.536,-32.768) | | (-49.152,-32.768) | 0 | |

Table 13: The second generation of $f_5$

As up implementation with GA, up time is need for finding global optimal point so GA will be combined with Parallel for accelerating speed up.

In these ways, the GA [12], called parallel genetic algorithms (PGAs), is commonly used to improve the convergence rate of GAs.

### 3.3 The Analysis and Design of Parallel Genetic Algorithm for SLM

Genetic Algorithms (GAs) [13] are a heuristic random search method widely used to find proper solutions to a variety of NP problems within a reasonable amount of time.

However, when they are applied to more complicated problems the time required to find an adequate solution increases. Therefore, past scholars have looked for multiple ways by which to accelerate the convergence rate of GAs.

There are three major types of PGAs: (1) the master-slave model, (2) island model and (3) fine-grained model. In the conventional master-slave model, one computing node is designated as the master while the others are designated a slaves. The master node holds the population and performs most of the GA operations. The master will assign one or more evolutionary tasks (i.e. fitness evaluation, crossover and mutation) to the slaves. This is done by sending one or more chromosomes to the slaves, after which the master will return the best result received from the slave nodes. In a coarse-grained model, the population is divided into several nodes. Each node then has a sub-population on which it executes GA operations. In a fine-grained model, each node only has a single chromosome, and each node can only communicate with several neighboring nodes.

In this case, the population is the collection of all the chromosomes in each node. To execute a genetic operation, a computing node must interact with its neighbors. Since the neighborhoods overlap, eventually the good characteristics of a superior individual can be distributed to the entire population. Fine-grained models have a large communication overhead due to the high frequency of interactions between neighboring nodes. More related details of PGA are introduced in [14]-[16], [13]. In this paper, we adopt the structure of the master-slave model as Figure 17 to achieve the objective of parallel processing.

Figure 17: The structure of the Adaptive Clustering-Based Parallel Genetic Algorithm with Migration (CBPGA)

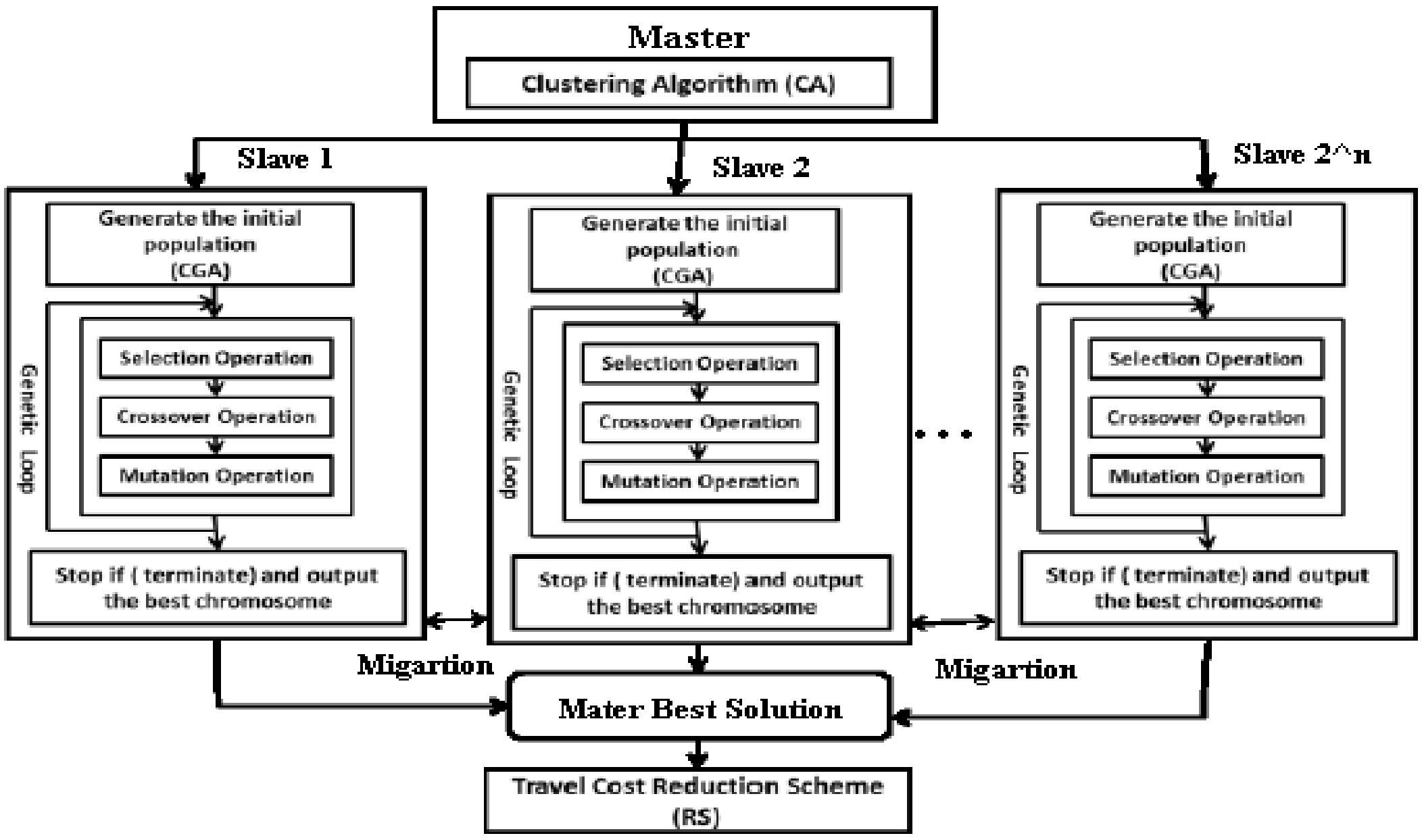

In SLM, when search space is divided, creating several other search spaces that the number of search space in different dimensional is shown table 14.
For example for 2- dimensional after subdividing, we have 4 search spaces like figure 19. Every search space has several nodes as table 14. At present, earning complete label in master is our aim. Every slave use GA operation (selection, mutation) in every search space then will produce label of every points in slave. For example figure19 that is for 2- dimensional and 4 points belong to one every search space, every search space belongs to one slave and every slave earn label of every points. At last, all of slaves send their solutions to master and master select the best solution that this is complete label .as figure 20, slave 3 has complete labeling. After this, search to global optimal point is in this search space and other search spaces are rejected.

| **Dimension** | **Search spaces** | **nodes** | **Processor** |
|---|---|---|---|
| 2 | 4 | 4 | 4 |
| 3 | 8 | 8 | 8 |
| ⋮ | ⋮ | ⋮ | ⋮ |
| n | $2^n$ | $2^n$ | $2^n$ |

Table 14: Search spaces in different dimensional

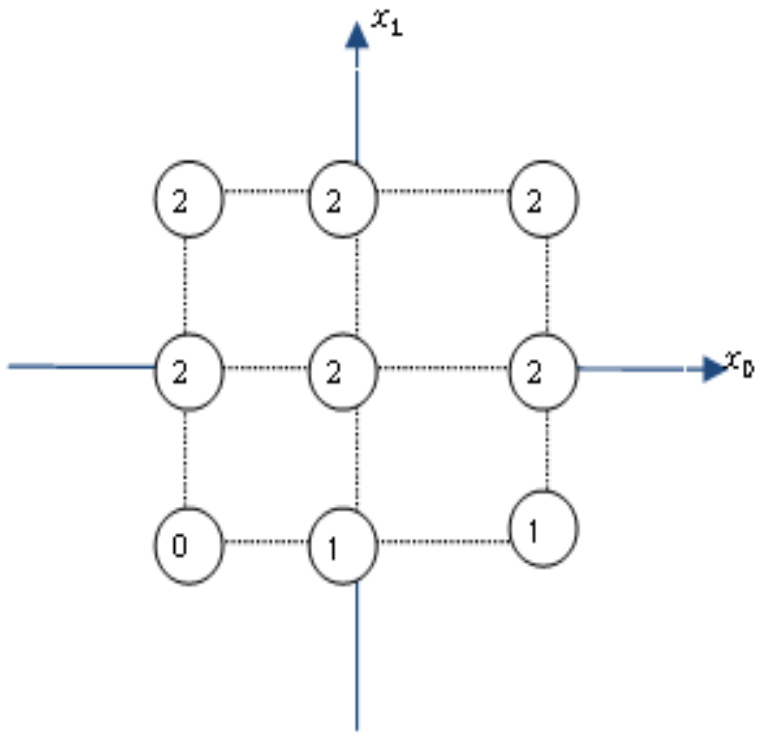

Figure 18: The number of search space in 2 -dimensional

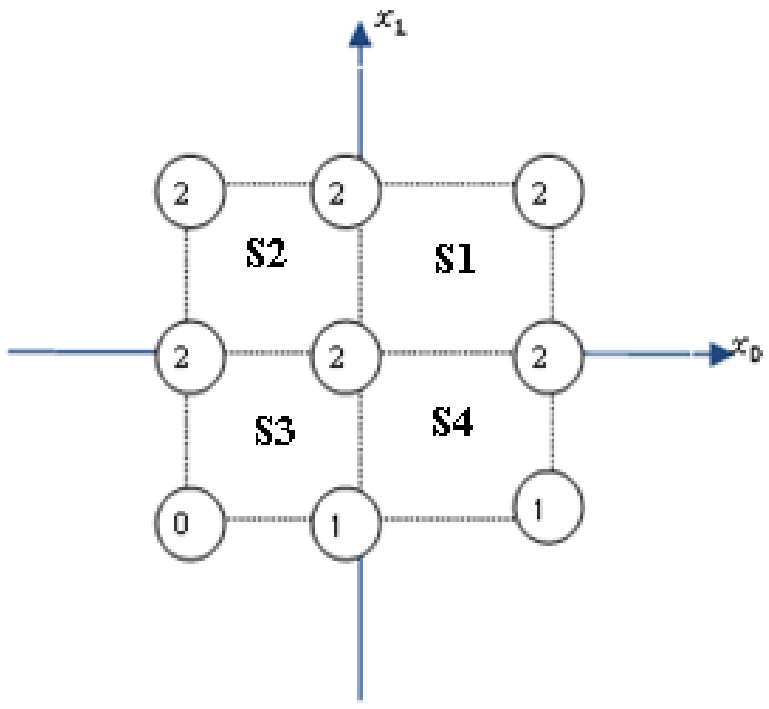

Figure 19: The number of slaves in 2 -dimensional

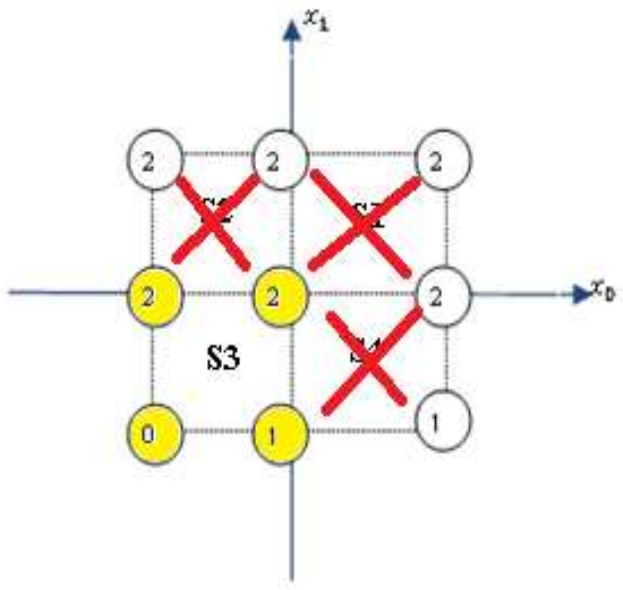

Figure 20: Earning slave has complete label

### 3.3.1 Schema of SLMPGA (SLMPGA) with Slave Master model

This schema is PLMGA with Master-Slave model for SLM.

Figure 21: SLMPGA with master _ slave model

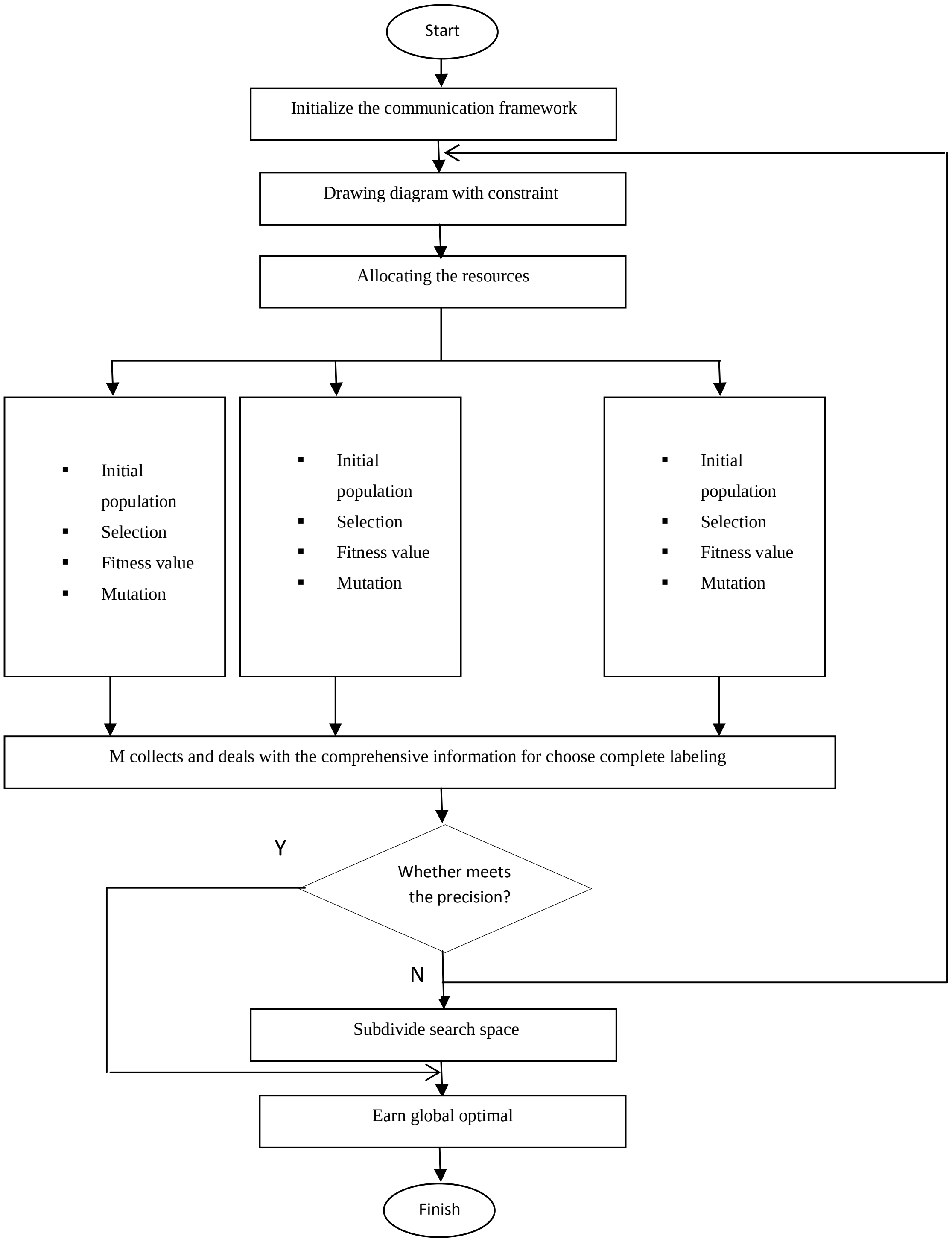

### 3.4 Clustering Algorithm in SLMPGA

Since optimization problem is a NP-hard problem [17], [18], and in order to accelerate the convergence rate of generating the near-optimal for reducing run time, and speed up .we combine the structure of the master-slave PGA (parallel genetic algorithm) [14]-[16] with the clustering algorithm.

As Schema of SLMPGA in step 3.3.1 , every slave was responsible for one search space and  every search space  had several points so points that are in one search space are given to one cluster and this is achieved through the clustering algorithm (CA) and when combined with SLMPGA , SCBPGA are achieved.

Here, as figure 22 and 23 the CA classifies several points that share a search space as the same cluster.  Every search space has $2^n$ -1 joint point to the other search spaces. Every search space need to all of information its points so we cluster points of exist in one search space and every point in one search space give their information to one cluster so every point can give its information to other clusters.

In introducing CA, two tables are created (show jointly as Table 15) to record the information of every points and clusters for quick reference with the clustering algorithm. The point- Cluster Table has two columns (points vs. Cluster ID),
Where each point is matched with its corresponding clusters and initial values are all set to null. The reciprocal is true of the Cluster-point Table.

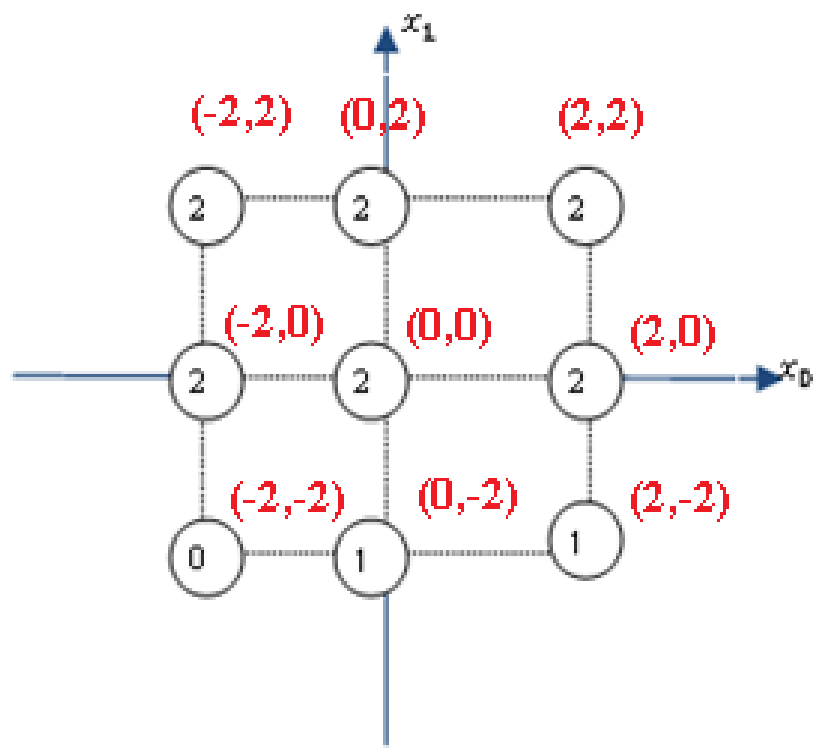

Figure 22: This figure shows the search space

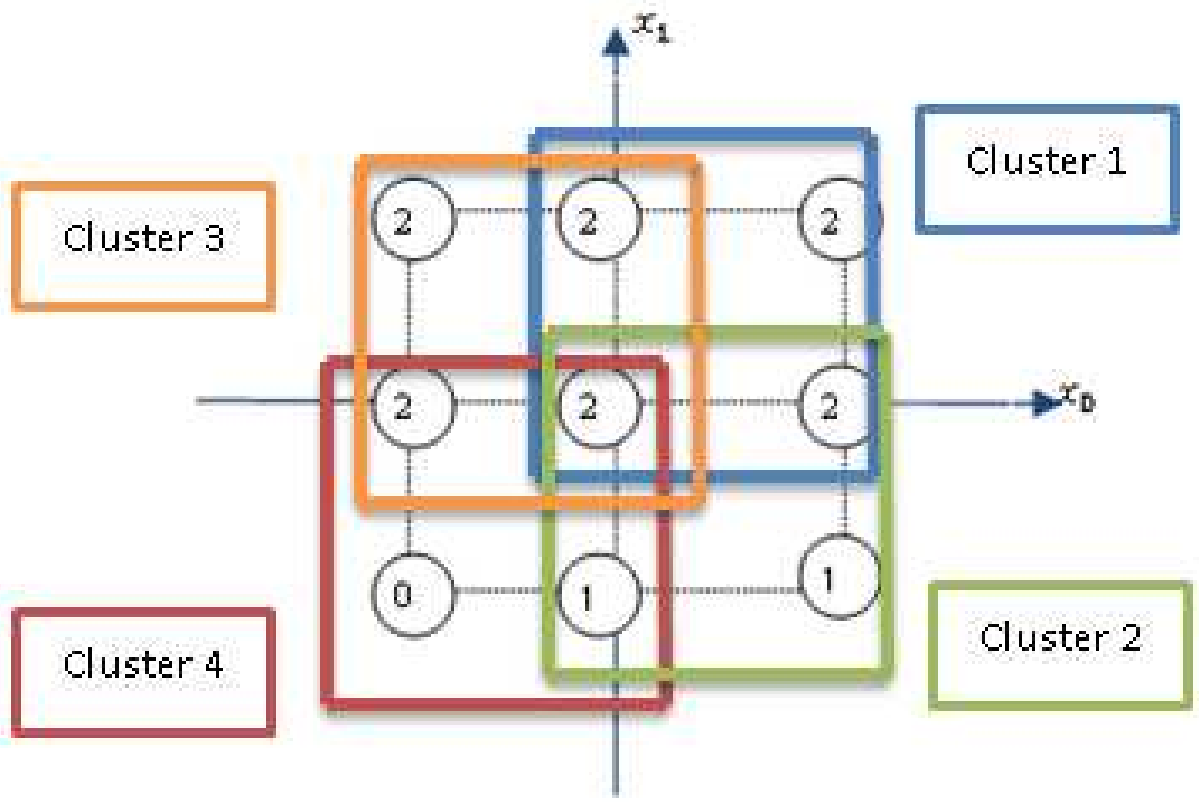

Figure 23: This figure shows four clusters.

| Point | Cluster ID |
|---|---|
| **(2,2)** | 1 |
| **(2,0)** | 1,2 |
| **(0,0)** | 1,2,3,4 |
| **(0,2)** | 1,3 |
| ⋮ | ⋮ |

| Cluster ID | Point |
|---|---|
| **1** | (2,2),(2,0),(0,0),(0,2) |
| **2** | (0,0),(2,0),(0,-2),(2,-2) |
| ⋮ | ⋮ |

Table 15: two tables for relationship between Points and clusters

### 3.5 SLMCBPGA

For CBPGA, GA in section 3.2, PGA in section 3.3 CA in section 3.4 are discussed.
The structure of CBPGA is depicted in figure 19, the main tasks of the CBPGA master node include executing the clustering algorithm, creating the initial population by CGA, calculating the fitness value for each chromosome and ensuring each slave shares the best chromosome information with each other. When a slave receives the population information from the master, the slave performs the whole evolutionary process (i.e. selection operation, mutation operation) until the terminal condition is satisfied. A slave only returns back with the best result when the evolutionary process is terminated. When the master receives the results from all slaves it selects the best result from these.

## 4 Experimental Setup

In this section, we implement SLMCBPGA, SLMPGA and SLMPA for De Jong Functions F2.in experiment is assigned each search space to one processor so the number of processors in different dimensional is different as table 14. We perform 30 trials of each CBPGA, SLMPGA and SPA on each problem and for each processor setting. Where P is the number of processors and d is the dimension of the problem. In all experimental and algorithms (SLMCBPGA, SLMPGA and SLMPA), we follow global minimum.
The algorithms are compared in terms of average time required to get a solution , where the global minimum value is not known, we measure time required to get a solution in which of algorithms have less time .
All numerical computations are performed in C++ using several 2.4 GHz Processor Pentium 4 machines times for SLMCBPGA, SLMPGA and SLMPA can be estimated reasonably well by running the parallel algorithms sequentially.

## 5 Experimental Results

| Algorithm | NP | Time | LB Time | Speedup | Efficiency |
|---|---|---|---|---|---|
| **SPA** | **1** | 40.485 | 40.485 | 1 | 1 |
| | **2** | 28.245 | 20.242 | 1.433 | 0.716 |
| | **3** | 28.840 | 13.495 | 1.403 | 0.467 |
| | **4** | 15.581 | 10.121 | 2.598 | 0.649 |
| | **5** | 15.781 | 8.097 | 2.565 | 0.513 |
| | **6** | 15.987 | 6.747 | 2.532 | 0.422 |
| | **7** | 16.210 | 5.783 | 2.497 | 0.356 |
| | **8** | 8.562 | 5.060 | 4.728 | 0.591 |
| | **9** | 8.569 | 4.498 | 4.724 | 0.524 |
| | **10** | 8.572 | 4.048 | 4.722 | 0.472 |
| | | | | | |
| **SPGA** | **1** | 33.215 | 33.215 | 1 | 1 |
| | **2** | 27.251 | 16.607 | 1.218 | 0.609 |
| | **3** | 17.235 | 11.071 | 1.927 | 0.642 |
| | **4** | 15.369 | 8.303 | 2.161 | 0.540 |
| | **5** | 13.569 | 6.643 | 2.447 | 0.489 |
| | **6** | 11.012 | 5.538 | 3.016 | 0.502 |
| | **7** | 11.023 | 4.745 | 3.013 | 0.430 |
| | **8** | 6.321 | 4.151 | 5.254 | 0.656 |
| | **9** | 6.451 | 3.690 | 5.148 | 0.572 |
| | **10** | 6.569 | 3.321 | 5.056 | 0.505 |
| | | | | | |
| **SCBPGA** | **1** | 25.215 | 25.215 | 1 | 1 |
| | **2** | 17.253 | 12.607 | 1.461 | 0.730 |
| | **3** | 12.321 | 8.405 | 2.046 | 0.682 |
| | **4** | 10.258 | 6.303 | 2.458 | 0.614 |
| | **5** | 10.123 | 5.043 | 2.490 | 0.498 |
| | **6** | 9.362 | 4.202 | 2.693 | 0.448 |
| | **7** | 8.369 | 3.602 | 3.012 | 0.430 |
| | **8** | 6.321 | 3.151 | 3.989 | 0.498 |
| | **9** | 6.123 | 2.801 | 4.118 | 0.457 |
| | **10** | 6.123 | 2.521 | 4.118 | 0.411 |

## F1 and SPA

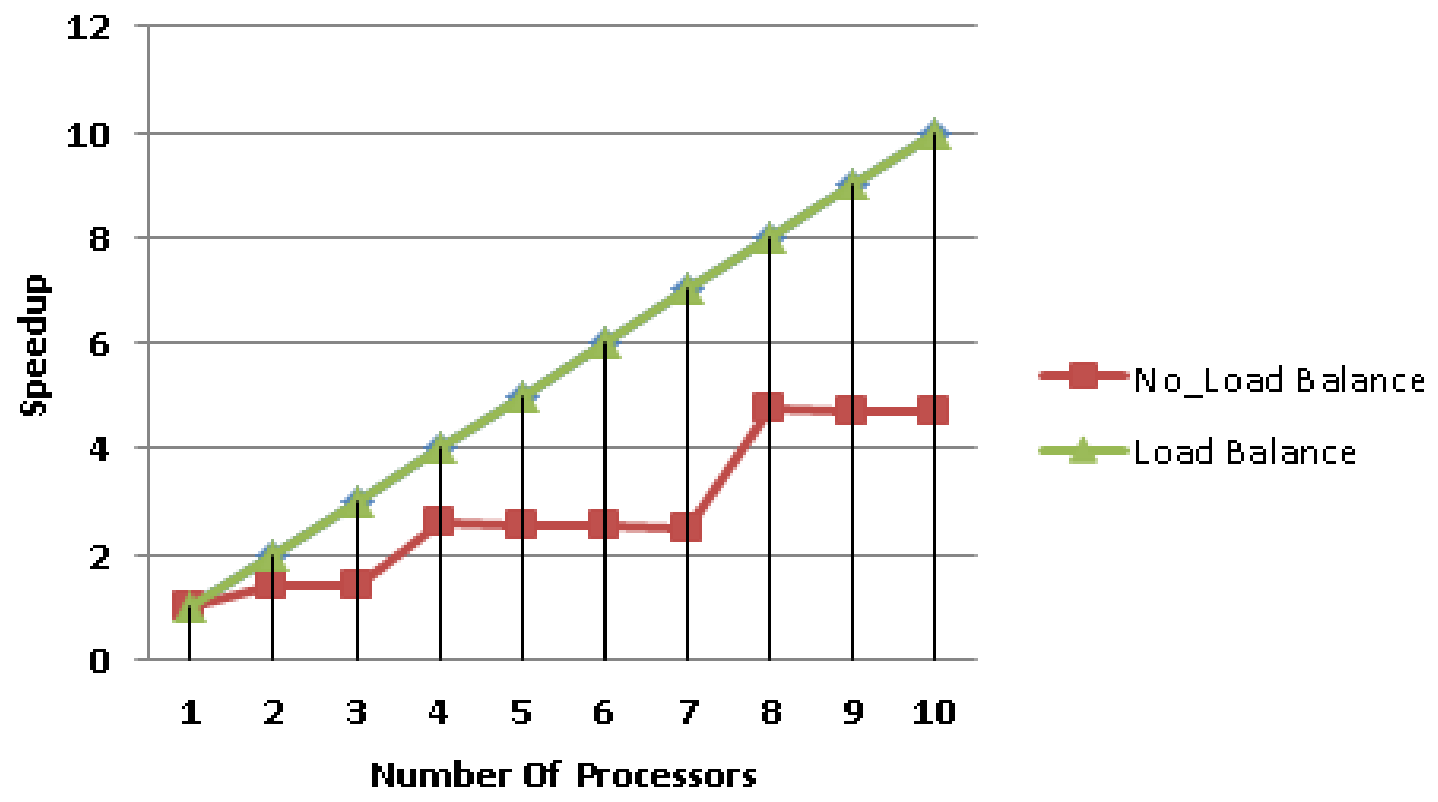

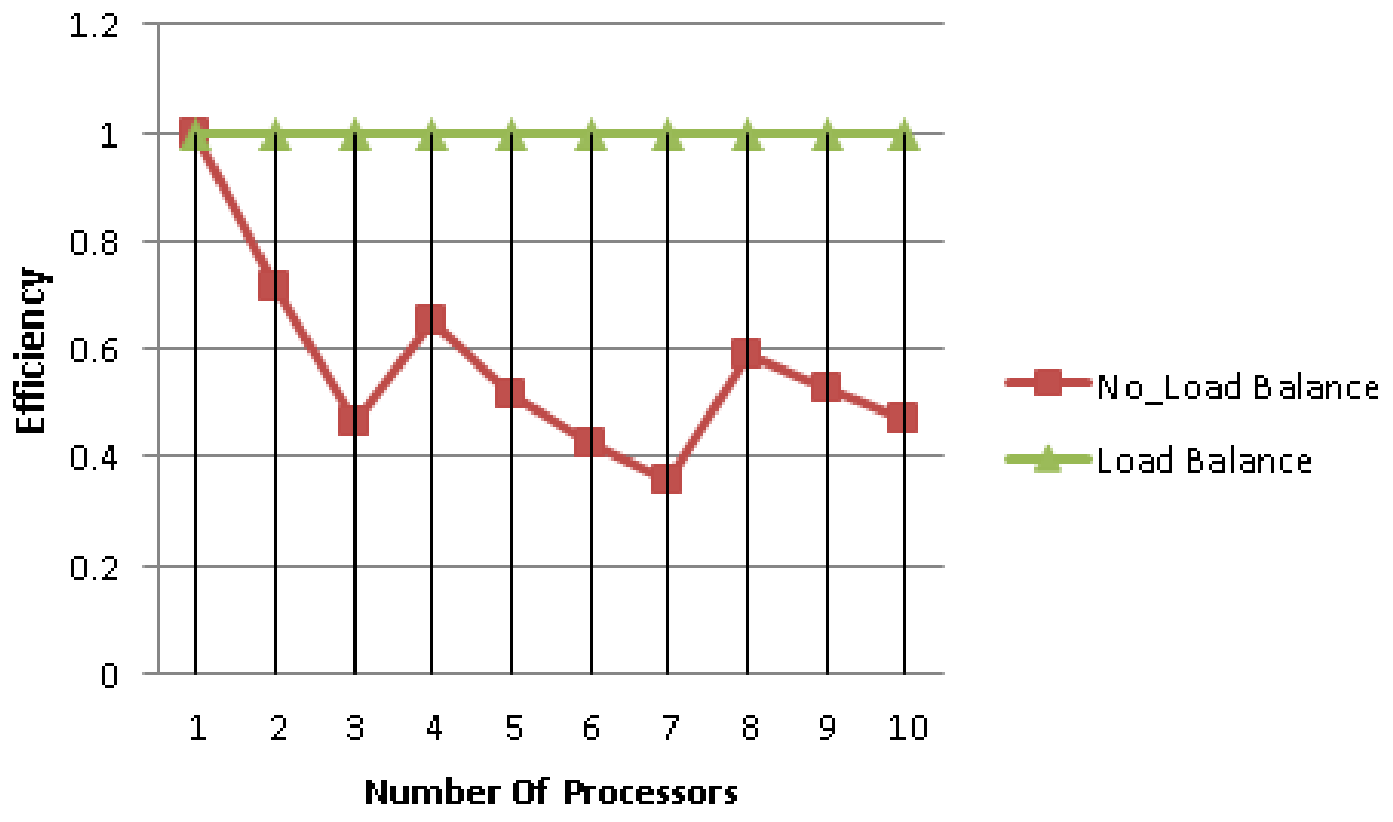

## F1 and SPGA

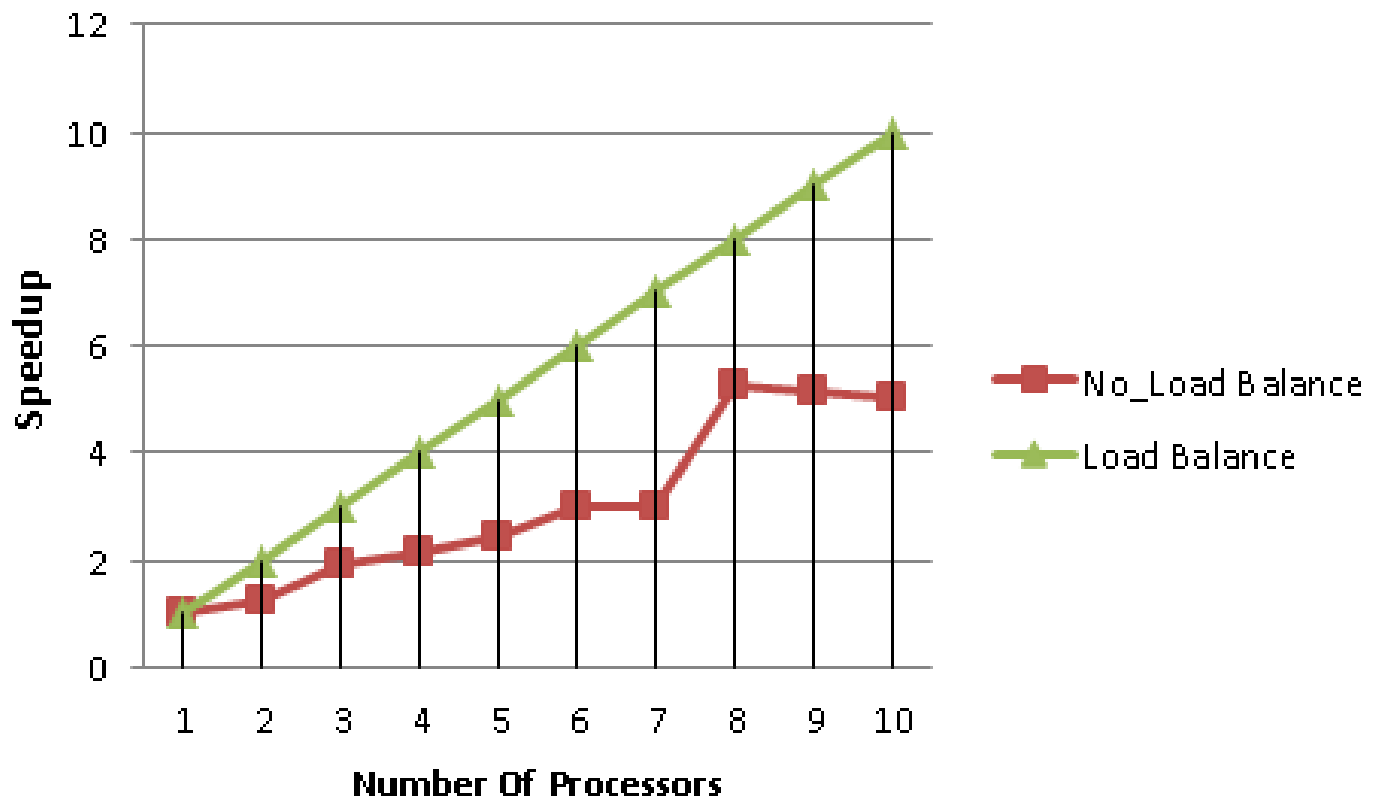

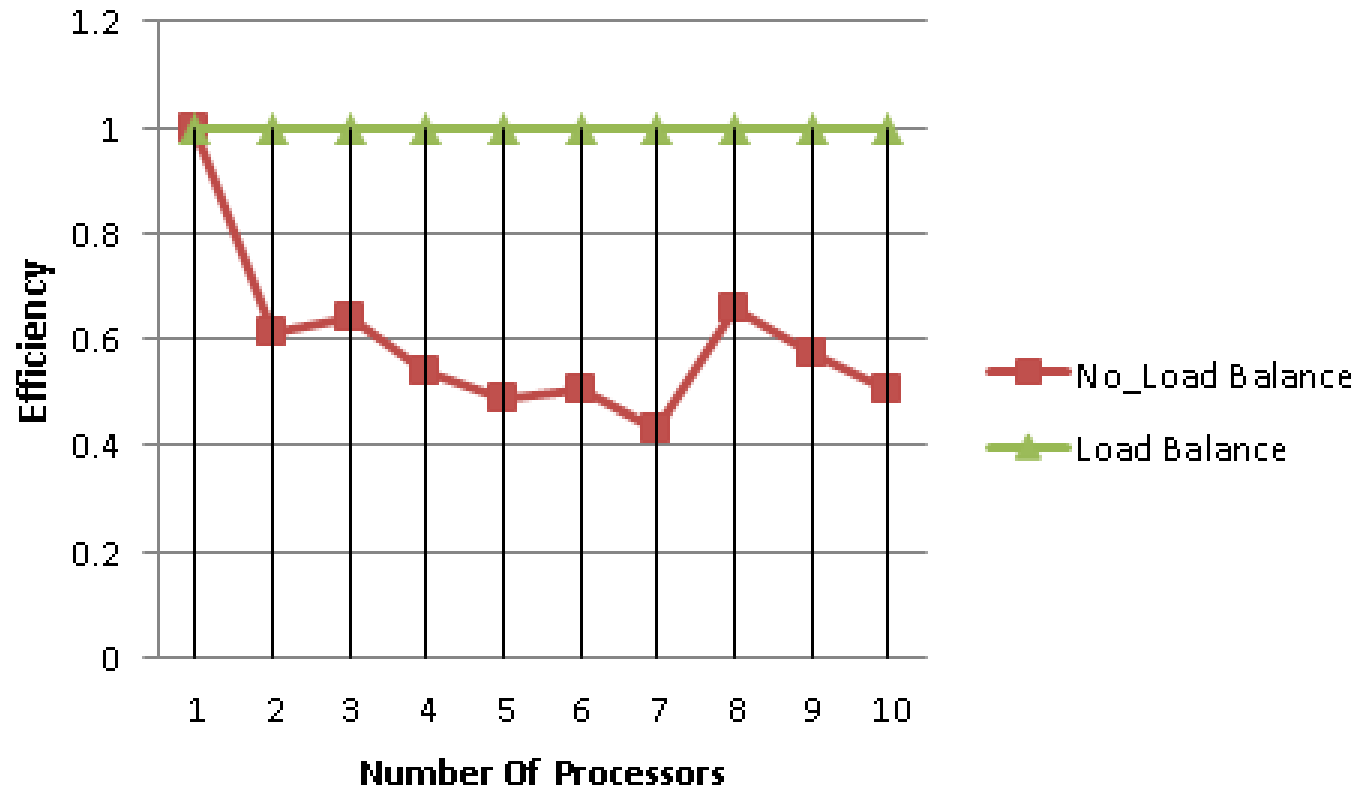

## F1 and SCBPGA

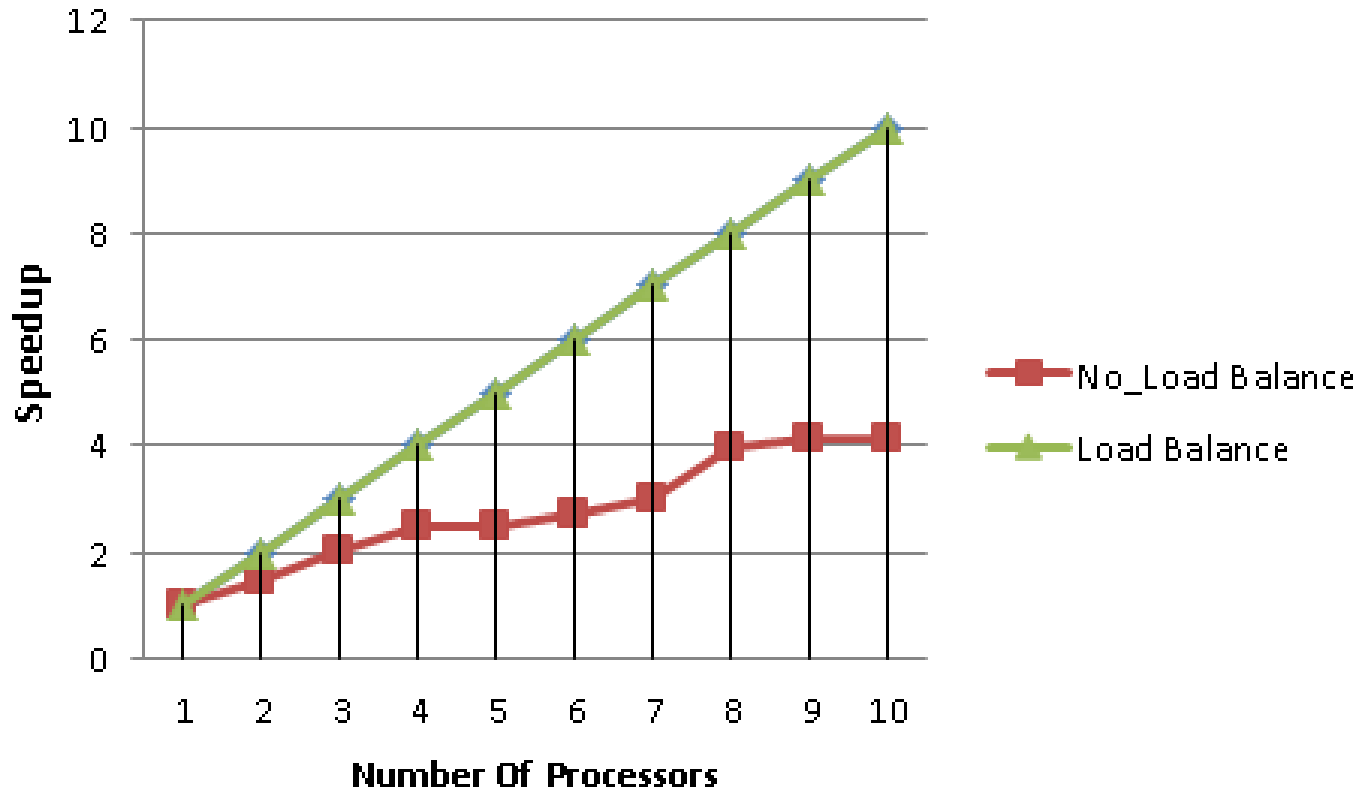

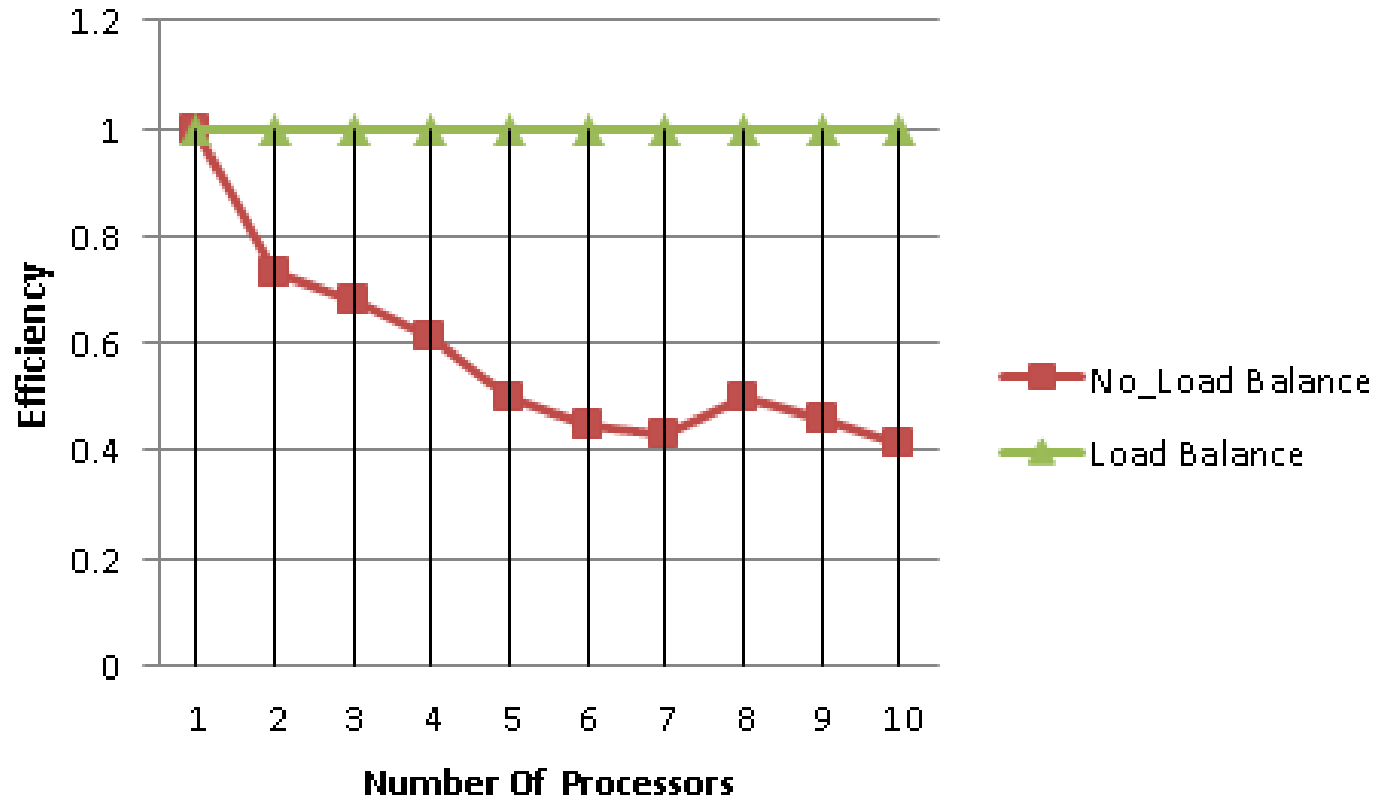

## 6 Conclusions

This paper is proposed a new method for finding global maximum and minimum point in optimization problems that is called SLM. As according to test problem and table 1,2, 3, SLM earn global point that the other methods earn with better time complexity .
Then is proposed a clustering-based parallel genetic algorithm (CBPGA) for solving the SLM. Based on the simulation results obtained, GA reduces the search space, parallelism accelerates convergence and cluster gathers information of joint points in different search space.
.Overall, the combined effects of clustering, parallelism and GA with CBPGA effectively and efficiently find the optimal point within fewer generations and higher speed than other evolutionary methods, such as simple SLM, GA and PGA.